\documentclass[pdflatex,sn-mathphys-num]{sn-jnl}

\usepackage{graphicx}%
\usepackage{multirow}%
\usepackage{amsmath,amssymb,amsfonts}%
\usepackage{amsthm}%
\usepackage{mathrsfs}%
\usepackage[title]{appendix}%
\usepackage{xcolor}%
\usepackage{textcomp}%
\usepackage{manyfoot}%
\usepackage{booktabs}%
\usepackage{algorithm}%
\usepackage{algorithmicx}%
\usepackage{algpseudocode}%
\usepackage{listings}%
\usepackage{float}
\usepackage{threeparttable}
\usepackage{changepage}
\usepackage{amsmath,amssymb,amsfonts}%
\usepackage{mathtools}%
\usepackage{array}
\usepackage{placeins}

\usepackage[table]{xcolor}
\usepackage{booktabs,multirow,makecell,pifont}
\usepackage{wrapfig} 

\newcommand{\cmark}{\textcolor[rgb]{0.800,0.133,0.133}{\ding{51}}}
\newcommand{\xmark}{\textcolor[rgb]{0.133,0.545,0.133}{\ding{55}}}
\newcommand{\rev}[1]{\textcolor{red}{#1}}
\newcommand{\StatexX}{\Statex\hspace{\algorithmicindent}}

\theoremstyle{thmstyleone}%
\theoremstyle{thmstyletwo}%

\theoremstyle{thmstylethree}%

\begin{document}

\title[Article Title]{RDMA: Cost Effective Agent-Driven Rare Disease \rev{Mining from Electronic Health Records}}


\author*[1]{\fnm{John} \sur{Wu}}\email{johnwu3@illinois.edu}
\author[1]{\fnm{Jimeng} \sur{Sun}}\email{jimeng@illinois.edu}
\author[2]{\fnm{Adam} \sur{Cross}}\email{arcross@uic.edu}

\affil*[1]{\orgdiv{Department of Computer Science}, \orgname{University of Illinois Urbana-Champaign}, \orgaddress{\city{Champaign}, \state{IL}, \country{USA}}}
\affil[2]{\orgdiv{Department of Pediatrics}, \orgname{University of Illinois College of Medicine Peoria}, \orgaddress{\city{Peoria}, \state{IL}, \country{USA}}}


\abstract{\rev{Rare diseases affect 1 in 10 Americans yet remain systematically underdocumented in clinical records. ICD-based systems cannot capture their breadth, over 50\% of Orphanet codes lack a direct ICD mapping and only 2.2\% of HPO codes have matching ICD codes, leaving patient populations invisible and delaying diagnosis. Mining unstructured clinical notes offers a direct path forward, but real notes are long, noisy, and abbreviation-dense, and limited annotations make fine-tuning infeasible, demanding approaches that generalize without task-specific training. We present Rare Disease Mining Agents (RDMA), an agentic framework equipping smaller quantized LLMs with tools for abbreviation resolution, implicit phenotype reasoning, and ontology grounding against Orphanet and HPO. RDMA substantially outperforms fine-tuned and RAG-based baselines across benchmarks with different data characteristics, without any task-specific training. A small quantized model achieves maximal performance, reducing inference costs by up to 10x and local hardware costs by up to 17x, enabling private deployment on standard hardware without cloud-based PHI exposure. RDMA's uncertainty-flagging mechanism further reduces expert annotation burden while preserving agreement quality, supporting scalable rare disease documentation in clinical practice. Available at https://github.com/jhnwu3/RDMA.}
}

\keywords{Rare Disease, Agents, Data Mining}



\maketitle
\section{Introduction} \label{ref:Intro}
Rare diseases affect approximately 1 in 10 Americans, constituting a significant healthcare challenge despite their individual scarcity \cite{VTNews2025_general_fact}. Accurate diagnosis remains difficult due to the vast diversity and sparsity of these conditions \cite{auvin2018problem_rarity}. While efforts to map \rev{international classification of diseases (ICD)} codes to more granular Orphanet codes have been attempted \cite{cavero2020icd10_to_orpha}, over 50\% of Orpha codes lack a direct mapping, resulting in the systematic under-reporting of rare diseases in ICD-annotated systems. Furthermore, only 2.2\% of HPO codes have matching ICD codes \cite{tan2024implications_icd_codes}. \rev{Existing rare disease differential diagnosis benchmarks rely directly on ICD labels without Orphanet or HPO concept annotations \cite{zhao2026agenticrarediseasediagnosis_deeprare}, potentially overstating real-world performance \cite{nguyen2026diagnostic_diff_diag_meta_analysis}. General purpose clinical NLP tools such as cTAKES \cite{wiegreffe2019clinical_ctakes} and MetaMap \cite{khin2017medical_metamap} inherit this same limitation, as they ground to ICD and UMLS terminologies, making automated documentation of rare disease conditions challenging.}
\rev{Mining clinical notes from EHRs offers a promising path forward, enabling direct mapping to HPO \cite{gargano2024mode_hpo} and Orphanet \cite{weinreich2008orphanet} without relying on ICD codes as intermediaries \cite{dong2023ontology_poor_annotations, yang2023enhancingphenotyperecognitionclinical}. However, real clinical notes are long, noisy, and filled with abbreviations and implicit clinical reasoning, exposing three critical gaps in existing mining approaches.}

\textbf{First, existing public mining benchmarks are either poorly annotated or fail to reflect real-world clinical notes, making fine-tuning on them unreliable or infeasible.} For example, our physicians found that annotations in earlier MIMIC-III rare disease mention mining attempts \cite{johnson2016mimic3, dong2023ontology_poor_annotations} frequently misinterpreted clinical abbreviations, such as interpreting NPH as normal pressure hydrocephalus'' rather than neutral protamine hagedorn'' in insulin treatment contexts. Additionally, phenotype mining approaches like RAG-HPO \cite{garcia2024_HPORAG} and PhenoGPT \cite{yang2023enhancingphenotyperecognitionclinical} evaluate on clinical case studies that lack the abbreviations and misspellings found in real clinical notes, and these case studies \cite{martinez2022raredis, yang2023enhancingphenotyperecognitionclinical} typically contain only several hundred words, whereas real clinical notes span thousands \cite{edin2023automated_medical_coding_survey, johnson2023mimic4_note}. Several studies \cite{wu2024hybrid, chen2024rarebenchllmsserverare} further rely on private annotations, impeding reproducible development. \rev{As a key consequence, approaches trained on one benchmark often fail to generalize across benchmarks with different data characteristics, and fine-tuning for specific deployment settings is frequently unrealistic, demanding methods that are robust without any task-specific training.}

\textbf{Second, a substantial portion of clinically relevant phenotypes remain implicit rather than explicitly stated in notes.} \rev{Identifying these requires interpreting laboratory results and applying clinical reasoning, tasks that go beyond simple entity extraction.} Conventional approaches typically follow a two-stage extract-and-match pipeline that fails to capture these implied phenotypes. Most existing methods \cite{garcia2024_HPORAG, wu2024hybrid, chen2024rarebenchllmsserverare} first extract entities and then match or verify them against ontologies like HPO \cite{gargano2024mode_hpo} or Orphanet \cite{weinreich2008orphanet} using retrieval-augmented generation (RAG) \cite{gao2024retrievalaugmentedgenerationlargelanguage_RAG_SURVEY}. Dictionary-based approaches \cite{deisseroth2019clinphen, liu2019doc2hpo, yadav2018survey_ner} like FastHPOCR \cite{groza2024fasthpocr}, while extremely efficient, lack the ability to reason about implied entities or the broader context surrounding identified mentions.

\textbf{Third, privacy concerns present significant barriers to deploying capable models in practice.} While cloud services increasingly adopt HIPAA compliance measures \cite{ayad2011addressing_hipaa_azure, keshetti2025designing_cloud_hipaa}, deploying LLM APIs with protected health information (PHI) typically requires extensive institutional review board (IRB) scrutiny \cite{grady2015institutional}. Locally-deployable solutions offer advantages in privacy protection and audit transparency \cite{sun2024activeprivacyauditingsupervised_whitebox_auditing}, but local deployment has historically required large, expensive hardware.

\rev{These three challenges together define what an effective mining solution must provide: generalization across data characteristics without task-specific training, reasoning capabilities that go beyond surface-level extraction, and practical deployability under real-world privacy constraints. Large language models (LLMs) have emerged as natural candidates given their flexibility, strong performance across clinical tasks \cite{wang2024surveylargelanguagemodels_biomed}, and ability to leverage existing knowledge bases such as HPO and Orphanet without extensive retraining \cite{garcia2024_HPORAG, sanmartin2024kg_rag, gao2024retrievalaugmentedgenerationlargelanguage_RAG_SURVEY}. LLMs can also provide interpretability through explanations \cite{savage2024diagnostic_interpretability}, making them well-suited for mining new phenotypes that require reasoning beyond semantic similarity. However, existing LLM-based approaches do not resolve the three challenges above: they are evaluated on private benchmarks \cite{wu2024hybrid, chen2024rarebenchllmsserverare}, rely on explicit entity extraction without reasoning about implied phenotypes \cite{garcia2024_HPORAG, wu2024hybrid}, and typically require cloud deployment of large models \cite{garcia2024_HPORAG}. Critically, we find that simply scaling to larger LLMs does not overcome these reasoning limitations, making the design of the agentic workflow, rather than model size, the determining factor in robust performance.}

\begin{figure}[htbp]
    \centering
    \includegraphics[width=1.0\textwidth]{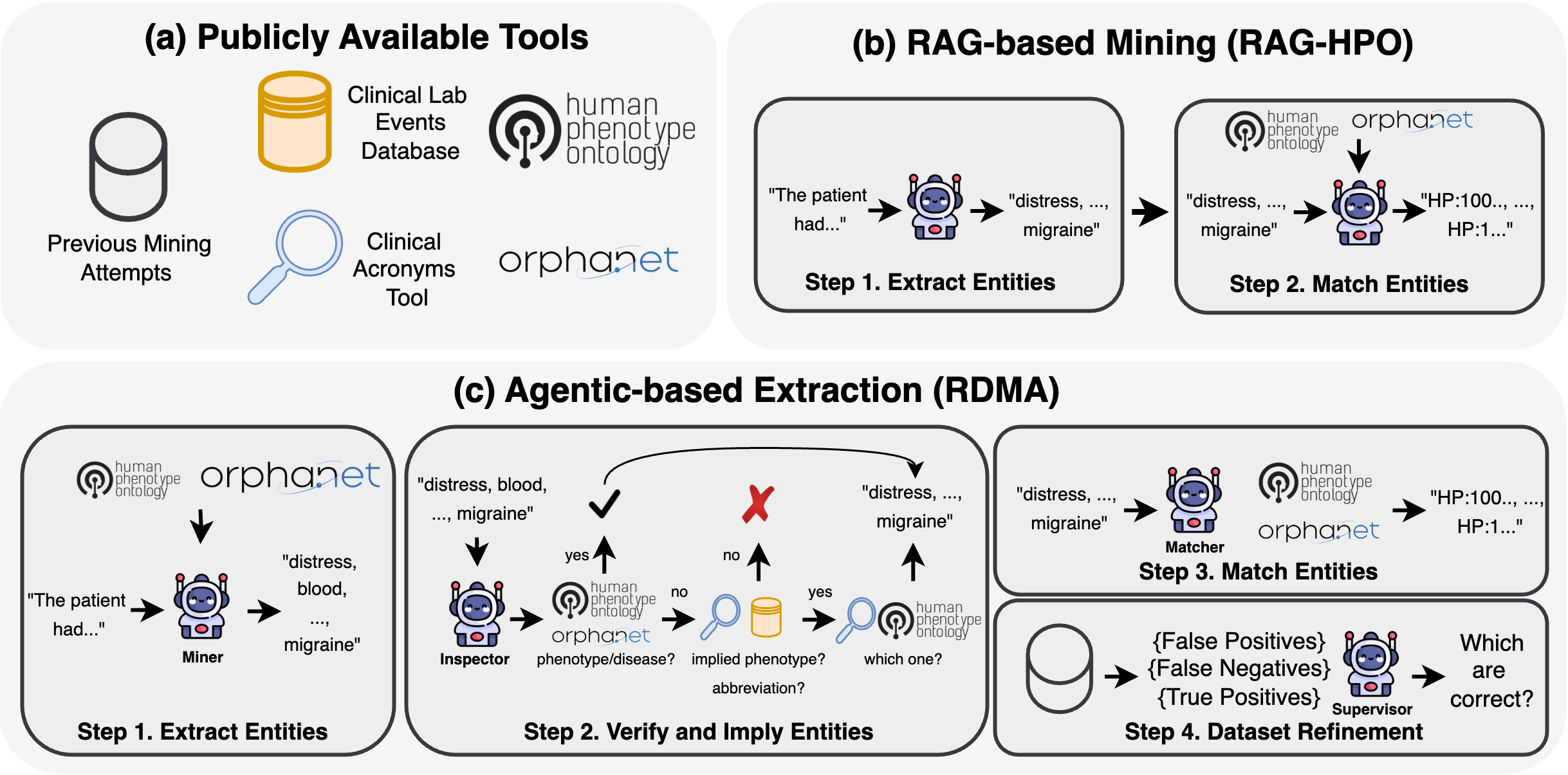}
    \caption{\textbf{RDMA vs. RAG-based approaches.} (b) RAG-based approaches \cite{garcia2024_HPORAG, wu2024hybrid, yang2023enhancingphenotyperecognitionclinical} typically follow a two-stage pipeline: first extracting entities using specialized extractors like SemEHR \cite{wu2024hybrid, wu2018semehr}, general-purpose LLMs \cite{garcia2024_HPORAG}, or finetuned LLMs \cite{yang2023enhancingphenotyperecognitionclinical}, then matching these entities to HPO codes \cite{wu2018semehr, wu2024hybrid, yang2023enhancingphenotyperecognitionclinical}. (c) Our RDMA approach extends this paradigm to an agentic framework \cite{wang2024survey_agents}, doubling the number of tools in our extraction pipeline, incorporating additional reasoning steps for verification and implication detection. Specifically, armed with lab event guidelines and a clinical acronyms database, the agent is forced to reason about whether an extracted entity implies a specific phenotype or rare disease. Furthermore, similar to how annotators re-review uncertain annotations, agents are required to perform a dataset refinement step to highlight challenging and uncertain cases \rev{by comparing previous mining attempts.}}
    \label{fig:RDMAvRAG}
\end{figure}

\rev{To address these challenges, we propose Rare Disease Mining Agents (RDMA), an agentic framework that coordinates LLM-based reasoning across structured ontology access, implicit phenotype inference, and uncertainty-flagging for human review (Fig.~\ref{fig:RDMAvRAG}). RDMA makes three key contributions: (1) it \textbf{generalizes substantially better across benchmarks without any task-specific training}, outperforming fine-tuned and RAG-based baselines across datasets with different data characteristics; (2) it \textbf{achieves maximal performance with a small 4-bit quantized model}, reducing inference costs by up to 10x and local hardware costs by up to 17x, making private deployment on standard consumer hardware (RTX 3090) practical; and (3) its \textbf{uncertainty-flagging mechanism reduces expert annotation burden} while preserving agreement quality, offering a blueprint for human-AI collaboration in refining noisy clinical datasets. We hope our exploration serves as a useful guide for building rigorous rare disease benchmarks and differential diagnosis frameworks.}

\section{Results}
Clinical note mining for rare disease differential diagnosis involves two key tasks: extracting phenotypes and identifying rare diseases. \rev{Here, we summarize the experimental setup towards evaluating existing mining approaches and RDMA.}

\textbf{Datasets.} \rev{Existing rare disease datasets have known limitations but remain useful benchmarks. \textbf{Phenotype identification.} We evaluate on the case study cohort (CSC) \cite{garcia2024_HPORAG} and BioLark gold standard corpora \cite{yang2023enhancingphenotyperecognitionclinical}. The CSC benchmark better reflects clinical notes with more text and phenotypes per document, though it still falls short of real-world clinical complexity. \textbf{Rare disease identification.} Three benchmarks exist: RDD \cite{fabregat2018deep_rdd}, RareDis \cite{martinez2022raredis}, and a noisily annotated MIMIC-III disease benchmark \cite{johnson2016mimic3,dong2023ontology_poor_annotations}. We exclude RDD as it lacks rare disease mention annotations. To demonstrate RDMA's ability to both improve extraction and assist annotators, we construct a gold evaluation set from the MIMIC-III annotations via combined human and agent annotation, yielding a realistic benchmark on real clinical notes. All benchmark statistics are described in Appendix \ref{appendix: cost calculations}.}

\rev{\textbf{Evaluation.} Our goal is to map text evidence to structured codes from the HPO \cite{gargano2024mode_hpo} and Orphanet \cite{weinreich2008orphanet} ontologies, framing extraction as a medical coding task \cite{Edin_2023_medical_coding_review}. \textbf{Phenotype extraction.} Annotated codes are available for both phenotype datasets, making evaluation straightforward. \textbf{Rare disease identification.} RareDis provides only text span annotations without ontology mappings. As both exist in our MIMIC3 evaluation set, we report both span and code mapping performance here as MIMIC3-RD Entity and MIMIC3-RD Code respectively. For baselines lacking an ontology mapping step, we choose to use a fuzzy matching retrieval approach that is employed by the authors of PhenoGPT \cite{yang2023enhancingphenotyperecognitionclinical}.} All evaluation metrics are described in Appendix \ref{appendix: Performance Metric Calculations}.

\begin{figure}[h]
    \centering
    \includegraphics[width=1.0\textwidth]{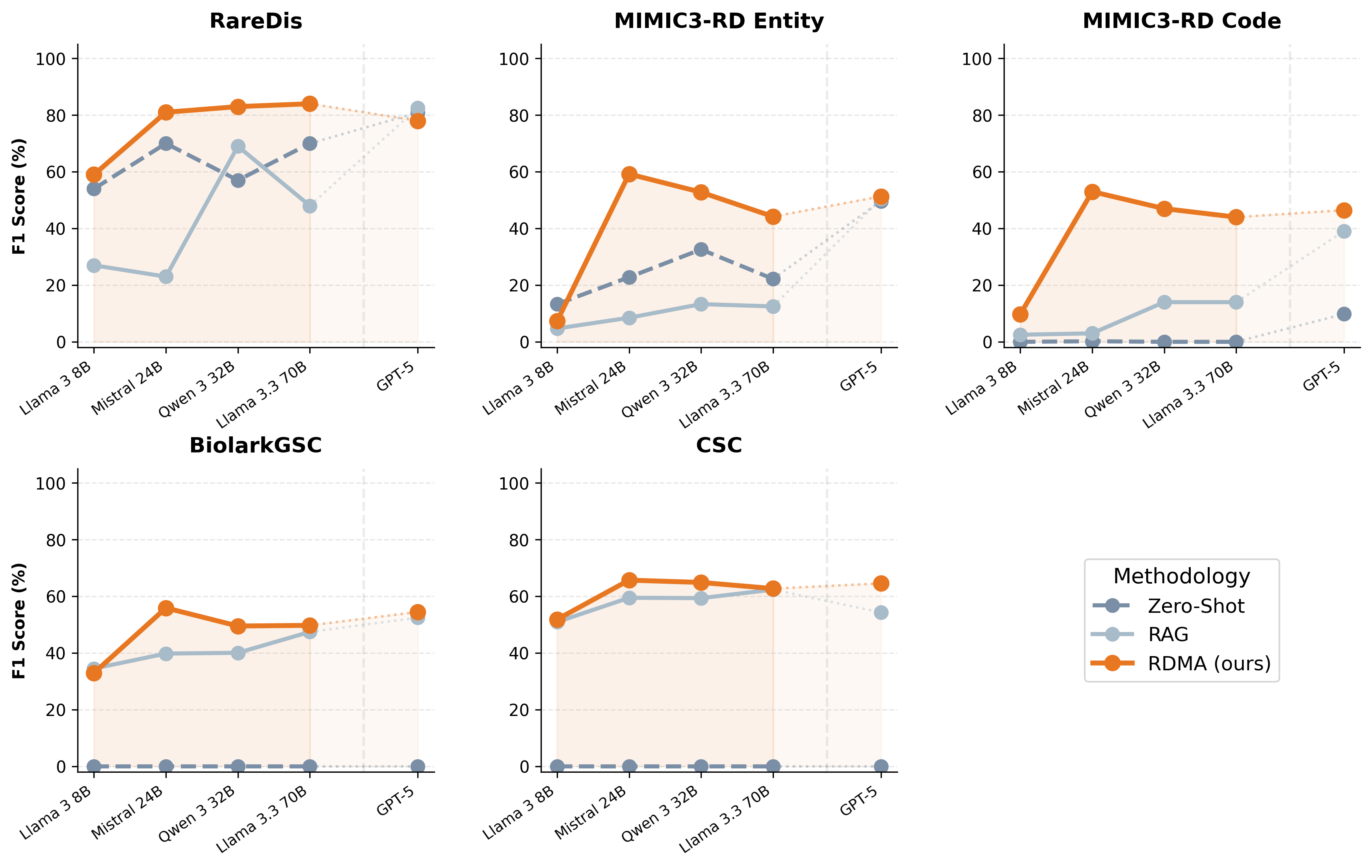}
    \caption{\rev{\textbf{Performance of LLM rare disease mining performance across model sizes and approaches.} We observe that in general increasing the model sizes does not improve performance of downstream mining approaches for rare diseases beyond a certain model size and intelligence, namely the Mistral 24B model. We also observe that mining performance for baseline RAG or zero-shot approaches fluctuate dramatically as seen on the RareDis and MIMIC3-RD Entity benchmarks. }}
    \label{fig:rd_model_perf_across_sizes}
\end{figure}

\textbf{Baselines.} \rev{Across all benchmarks, approaches fall into three categories: (1) \textbf{rule-based} methods that tokenize text and match to ontologies via nearest-neighbor search, (2) \textbf{fine-tuned} encoder models trained to extract entities directly, and (3) \textbf{LLM-based} methods using zero-shot or RAG-based extract-and-match pipelines.
For rule-based approaches, we include a simple retrieve-and-string-match \textbf{Dictionary Match} baseline using state-of-the-art clinical embeddings MedEmbed \cite{balachandran2024medembed} and the state-of-the-art \textbf{FastHPOCR}. For fine-tuned encoders, we distinguish between off-the-shelf models and those we fine-tune ourselves. Off-the-shelf, we evaluate \textbf{PhenoGPT} \cite{yang2023enhancingphenotyperecognitionclinical} and \textbf{Stanza's i2b2 Clinical BERT}, both used without modification. Following the RareDis authors \cite{segura2022exploring_raredis}, we additionally fine-tune \textbf{BioBERT} \cite{sun2021biomedical_biobert_ner} and \textbf{BioClinicalBERT} \cite{alsentzer2019publiclyavailableclinicalbert} for rare disease and phenotype extraction; all four encoder models are evaluated across all benchmark datasets. For LLM-based approaches, we test a range of models in both zero-shot and \textbf{RAG-based} settings \cite{garcia2024_HPORAG}. We note that highly specialized approaches such as FastHPOCR and PhenoGPT do not apply to our rare disease extraction benchmarks.}
\rev{\subsection{Extraction Performance}} \label{sec: Extraction Performance}
\rev{\textbf{We evaluate mining performance along two axes: generalization across distributions and scaling across model sizes in Table \ref{tab:main_results_combined_v3} and Fig. \ref{fig:rd_model_perf_across_sizes}.} Specifically, we ask (1) how well does each approach generalize across benchmarks with different data characteristics, and (2) how does performance scale with model size. We find that (1) our agentic system is substantially more robust than competing approaches, outperforming specialized baselines without requiring any training, and (2) a small quantized model is sufficient for maximal performance across all benchmarks, making RDMA computationally feasible compared to LLM-based approaches that depend on larger models for robust performance.}

\begin{table}[h]
  \centering
  \small
  \setlength{\tabcolsep}{3.5pt}
  \renewcommand{\arraystretch}{1.18}
  \resizebox{\textwidth}{!}{%
  \begin{tabular}{l c ccc ccc ccc ccc ccc}
    \toprule
    & & \multicolumn{6}{c}{\textbf{Phenotype Mining}} & \multicolumn{9}{c}{\textbf{Rare Disease Mining}} \\
    \cmidrule(lr){3-8} \cmidrule(lr){9-17}
    & & \multicolumn{3}{c}{\textbf{BiolarkGSC+}} & \multicolumn{3}{c}{\textbf{CSC}} & \multicolumn{3}{c}{\textbf{RareDis}} & \multicolumn{3}{c}{\textbf{MIMIC3-RD Entity}} & \multicolumn{3}{c}{\textbf{MIMIC3-RD Code}} \\
    \cmidrule(lr){3-5} \cmidrule(lr){6-8} \cmidrule(lr){9-11} \cmidrule(lr){12-14} \cmidrule(lr){15-17}
    \textbf{Approach} & \textbf{FT} & \textbf{P} & \textbf{R} & \textbf{F1} & \textbf{P} & \textbf{R} & \textbf{F1} & \textbf{P} & \textbf{R} & \textbf{F1} & \textbf{P} & \textbf{R} & \textbf{F1} & \textbf{P} & \textbf{R} & \textbf{F1} \\
    \midrule

    \multicolumn{17}{l}{\textit{Non-LLM Baselines}} \\[1pt]
    Dictionary Match & {\color{green}\xmark} & \cellcolor[rgb]{1.000,0.574,0.054}0.682 & \cellcolor[rgb]{1.000,0.829,0.620}0.214 & \cellcolor[rgb]{1.000,0.757,0.459}0.326 & \cellcolor[rgb]{1.000,0.599,0.110}0.600 & \cellcolor[rgb]{1.000,0.859,0.687}0.210 & \cellcolor[rgb]{1.000,0.788,0.528}0.310 & \cellcolor[rgb]{1.000,0.579,0.065}0.850 & \cellcolor[rgb]{1.000,1.000,1.000}0.410 & \cellcolor[rgb]{1.000,0.752,0.448}0.550 & \cellcolor[rgb]{1.000,0.653,0.229}0.400 & \cellcolor[rgb]{1.000,1.000,1.000}0.468 & \cellcolor[rgb]{1.000,0.677,0.281}\underline{0.431} & \cellcolor[rgb]{1.000,0.707,0.349}0.300 & \cellcolor[rgb]{1.000,0.690,0.312}0.450 & \cellcolor[rgb]{1.000,0.695,0.322}\underline{0.360} \\

    FastHPOCR & {\color{green}\xmark} & \cellcolor[rgb]{1.000,0.550,0.000}0.721 & \cellcolor[rgb]{1.000,0.586,0.080}0.518 & \cellcolor[rgb]{1.000,0.550,0.000}\textbf{0.603} & \cellcolor[rgb]{1.000,0.653,0.228}0.520 & \cellcolor[rgb]{1.000,0.698,0.329}0.450 & \cellcolor[rgb]{1.000,0.671,0.269}0.480 & \multicolumn{3}{c}{\cellcolor[gray]{0.95}\textcolor{red!80}{\textit{N/A}}} & \multicolumn{3}{c}{\cellcolor[gray]{0.95}\textcolor{red!80}{\textit{N/A}}} & \multicolumn{3}{c}{\cellcolor[gray]{0.95}\textcolor{red!80}{\textit{N/A}}} \\

    i2b2 Clinical BERT & {\color{red}\cmark} & \cellcolor[rgb]{1.000,0.626,0.169}0.599 & \cellcolor[rgb]{1.000,0.667,0.259}0.417 & \cellcolor[rgb]{1.000,0.634,0.186}0.491 & \cellcolor[rgb]{1.000,0.680,0.288}0.480 & \cellcolor[rgb]{1.000,0.598,0.106}0.600 & \cellcolor[rgb]{1.000,0.637,0.193}0.530 & \cellcolor[rgb]{1.000,0.987,0.970}0.153 & \cellcolor[rgb]{1.000,0.588,0.083}0.850 & \cellcolor[rgb]{1.000,0.977,0.948}0.260 & \cellcolor[rgb]{1.000,1.000,1.000}0.010 & \cellcolor[rgb]{1.000,0.550,0.000}0.879 & \cellcolor[rgb]{1.000,1.000,1.000}0.020 & \cellcolor[rgb]{1.000,0.991,0.980}0.010 & \cellcolor[rgb]{1.000,0.550,0.000}0.651 & \cellcolor[rgb]{1.000,0.986,0.968}0.019 \\

    BioBERT$^\dagger$ & {\color{red}\cmark} & \cellcolor[rgb]{1.000,0.604,0.119}0.635 & \cellcolor[rgb]{1.000,0.673,0.274}0.409 & \cellcolor[rgb]{1.000,0.628,0.174}0.498 & \cellcolor[rgb]{1.000,0.590,0.089}0.614 & \cellcolor[rgb]{1.000,0.814,0.586}0.278 & \cellcolor[rgb]{1.000,0.738,0.419}0.382 & \cellcolor[rgb]{1.000,0.627,0.171}0.730 & \cellcolor[rgb]{1.000,0.653,0.228}0.703 & \cellcolor[rgb]{1.000,0.620,0.153}\underline{0.716} & \cellcolor[rgb]{1.000,0.993,0.984}0.018 & \cellcolor[rgb]{1.000,0.900,0.779}0.559 & \cellcolor[rgb]{1.000,0.990,0.977}0.033 & \cellcolor[rgb]{1.000,0.976,0.948}0.025 & \cellcolor[rgb]{1.000,0.674,0.276}0.473 & \cellcolor[rgb]{1.000,0.962,0.915}0.047 \\

    BioClinicalBERT$^\dagger$ & {\color{red}\cmark} & \cellcolor[rgb]{1.000,0.714,0.363}0.459 & \cellcolor[rgb]{1.000,0.661,0.247}0.424 & \cellcolor[rgb]{1.000,0.671,0.269}0.441 & \cellcolor[rgb]{1.000,0.700,0.334}0.449 & \cellcolor[rgb]{1.000,0.767,0.481}0.348 & \cellcolor[rgb]{1.000,0.732,0.403}0.392 & \cellcolor[rgb]{1.000,0.650,0.233}0.681 & \cellcolor[rgb]{1.000,0.690,0.314}0.666 & \cellcolor[rgb]{1.000,0.660,0.255}0.673 & \cellcolor[rgb]{1.000,0.996,0.990}0.015 & \cellcolor[rgb]{1.000,0.995,0.988}0.473 & \cellcolor[rgb]{1.000,0.994,0.988}0.027 & \cellcolor[rgb]{1.000,0.975,0.946}0.026 & \cellcolor[rgb]{1.000,0.637,0.193}0.527 & \cellcolor[rgb]{1.000,0.959,0.909}0.050 \\

    \midrule
    \multicolumn{17}{l}{\textit{LLM-based Approaches}} \\[1pt]

    PhenoGPT & {\color{red}\cmark} & \cellcolor[rgb]{1.000,0.578,0.062}0.676 & \cellcolor[rgb]{1.000,0.751,0.446}0.312 & \cellcolor[rgb]{1.000,0.681,0.292}0.427 & \cellcolor[rgb]{1.000,0.619,0.154}0.570 & \cellcolor[rgb]{1.000,0.738,0.419}0.390 & \cellcolor[rgb]{1.000,0.685,0.300}0.460 & \multicolumn{3}{c}{\cellcolor[gray]{0.95}\textcolor{red!80}{\textit{N/A}}} & \multicolumn{3}{c}{\cellcolor[gray]{0.95}\textcolor{red!80}{\textit{N/A}}} & \multicolumn{3}{c}{\cellcolor[gray]{0.95}\textcolor{red!80}{\textit{N/A}}} \\

    Zero-Shot (Llama 3.3 70B) & {\color{green}\xmark} & \cellcolor[rgb]{1.000,1.000,1.000}0.000 & \cellcolor[rgb]{1.000,1.000,1.000}0.000 & \cellcolor[rgb]{1.000,1.000,1.000}0.000 & \cellcolor[rgb]{1.000,1.000,1.000}0.000 & \cellcolor[rgb]{1.000,1.000,1.000}0.000 & \cellcolor[rgb]{1.000,1.000,1.000}0.000 & \cellcolor[rgb]{1.000,0.550,0.000}0.900 & \cellcolor[rgb]{1.000,0.850,0.667}0.570 & \cellcolor[rgb]{1.000,0.635,0.190}0.700 & \cellcolor[rgb]{1.000,0.883,0.741}0.141 & \cellcolor[rgb]{1.000,0.853,0.674}0.602 & \cellcolor[rgb]{1.000,0.836,0.636}0.228 & \cellcolor[rgb]{1.000,1.000,1.000}0.001 & \cellcolor[rgb]{1.000,1.000,1.000}0.007 & \cellcolor[rgb]{1.000,1.000,1.000}0.002 \\

    RAG (RD / HPO) (Llama 3.3 70B) & {\color{green}\xmark} & \cellcolor[rgb]{1.000,0.744,0.431}0.410 & \cellcolor[rgb]{1.000,0.550,0.000}0.563 & \cellcolor[rgb]{1.000,0.646,0.212}0.475 & \cellcolor[rgb]{1.000,0.550,0.000}0.674 & \cellcolor[rgb]{1.000,0.611,0.136}0.580 & \cellcolor[rgb]{1.000,0.573,0.050}\underline{0.624} & \cellcolor[rgb]{1.000,1.000,1.000}0.130 & \cellcolor[rgb]{1.000,0.550,0.000}0.890 & \cellcolor[rgb]{1.000,1.000,1.000}0.230 & \cellcolor[rgb]{1.000,0.969,0.931}0.045 & \cellcolor[rgb]{1.000,0.728,0.397}0.716 & \cellcolor[rgb]{1.000,0.949,0.886}0.085 & \cellcolor[rgb]{1.000,0.984,0.965}0.017 & \cellcolor[rgb]{1.000,0.607,0.126}0.570 & \cellcolor[rgb]{1.000,0.976,0.947}0.030 \\

    \textbf{RDMA (Mistral 24B)}$^{\ast}$ & {\color{green}\xmark} & \cellcolor[rgb]{1.000,0.647,0.216}0.565 & \cellcolor[rgb]{1.000,0.558,0.018}0.553 & \cellcolor[rgb]{1.000,0.583,0.073}\underline{0.559} & \cellcolor[rgb]{1.000,0.570,0.045}0.644 & \cellcolor[rgb]{1.000,0.550,0.000}0.671 & \cellcolor[rgb]{1.000,0.550,0.000}\textbf{0.657} & \cellcolor[rgb]{1.000,0.568,0.039}0.870 & \cellcolor[rgb]{1.000,0.672,0.271}0.760 & \cellcolor[rgb]{1.000,0.550,0.000}\textbf{0.810} & \cellcolor[rgb]{1.000,0.550,0.000}0.516 & \cellcolor[rgb]{1.000,0.751,0.448}0.695 & \cellcolor[rgb]{1.000,0.550,0.000}\textbf{0.592} & \cellcolor[rgb]{1.000,0.550,0.000}0.460 & \cellcolor[rgb]{1.000,0.572,0.048}0.620 & \cellcolor[rgb]{1.000,0.550,0.000}\textbf{0.530} \\

    \bottomrule
  \end{tabular}%
  }
  \caption{\rev{\textbf{Standardized Performance Comparison Across Phenotype and Rare Disease Mining Approaches.} We evaluate both domains side-by-side using various LLM backbones for generative methods. For Rare Disease evaluation, BioBERT and BioClinicalBERT$^\dagger$ are fine-tuned on datasets (BiolarkGSC+, RareDis) where span annotations exist and evaluated on their evaluation counterparts (CSC, MIMIC3-RD) here. ``N/A" blocks denote models whose architectures or pre-training are not applicable to the respective dataset. Colors are normalized per benchmark so that the darkest orange always reflects the highest score within each column group. FT indicates whether the model was fine-tuned on any biomedical corpus before usage. \textbf{Bold} denotes best performance, \underline{underline} denotes second best. We report our bootstrapped confidence intervals in Appendix section \ref{appdx: CIs}. }}
  \label{tab:main_results_combined_v3}
\end{table}

\rev{\textbf{Agentic LLM approaches better generalize across distributions without domain-specific training.} For datasets where annotations do not contain explicit text spans and sample sizes are small, such as CSC and MIMIC3, fine-tuning is practically infeasible, and even where it is feasible, supervised models trained on one corpus frequently fail to generalize to out-of-distribution data. As shown in Table~\ref{tab:main_results_combined_v3}, fine-tuned models such as PhenoGPT, BioBERT, and BioClinicalBERT require extensive training on well-annotated corpora such as BiolarkGSC+ and RareDis, yet exhibit substantial performance degradation when evaluated on out-of-distribution benchmarks like CSC and MIMIC3. Rule-based approaches like FastHPOCR avoid this training requirement but exhibit similar brittleness: FastHPOCR \cite{groza2024fasthpocr} well on the benchmark it was originally evaluated on, but underperforms on CSC \cite{garcia2024_HPORAG}. By contrast, RDMA generalizes across extraction tasks without any additional training, providing a substantial performance uplift across 4 of the 5 benchmarks compared to both trained and rule-based alternatives.}

\rev{\textbf{LLMs hallucinate HPO and Orpha codes.} In benchmarks requiring both entity extraction and ontology code assignment, we observe that LLMs frequently hallucinate invalid HPO and Orpha codes despite correctly identifying the underlying text spans, as evidenced by the performance gap between text-span and code-based benchmarks in Table~\ref{tab:main_results_combined_v3}. Providing retrieved ontology context in the RAG setting substantially improves code-level performance, but often at the cost of recall as false positives increase. We hypothesize that retrieved ontology context biases the model toward over-predicting entity presence, surfacing mentions that are not actually attested in the source text.}

\rev{\textbf{Larger models do not improve extraction performance.} Beyond 8B models, which frequently struggle to follow structured instructions, we observe that increasing model size from 24B up to proprietary models such as GPT-5 yields no consistent improvement in rare disease or phenotype extraction performance across any benchmark, with gains effectively plateauing.}

\begin{wrapfigure}{r}{0.5\textwidth}
\centering
\includegraphics[width=0.5\textwidth]{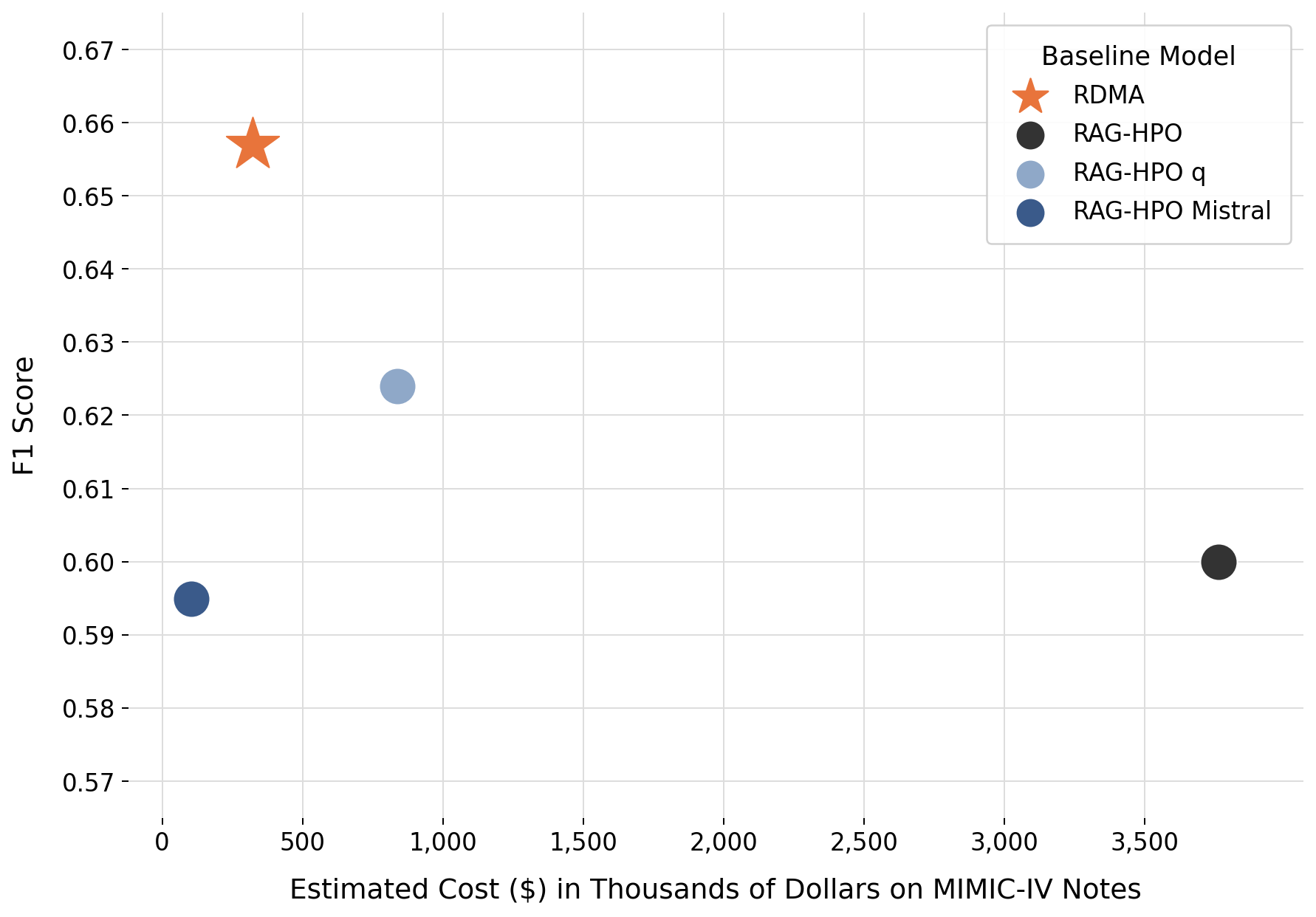}
\caption{\rev{\textbf{Phenotype mining inference costs with LLMs.} RDMA uses a 4-bit quantized Mistral 24B model yet outperforms RAG-HPO variants running on both a 4-bit and full non-quantized Llama 3.3 70B, models that are substantially more expensive to run locally. Although RDMA's additional reasoning steps incur a modest overhead when normalized by model size, this cost is far smaller than the penalty of scaling to larger models.} We use GPU rental pricing from cloud-providers \cite{salad_pricing} \cite{hyperstack_gpu_pricing} to compute our estimated inference costs in Appendix \ref{appendix: cost calculations}.}
\label{fig:performance_cost}
\end{wrapfigure}

\rev{\textbf{RDMA provides a more a stable foundation for LLM mining performance.} Different LLMs can respond very differently to the same prompts, and as shown in Fig.~\ref{fig:rd_model_perf_across_sizes}, extraction performance varies substantially across model families in both zero-shot and RAG settings on the RareDis and MIMIC3-RD Entity benchmarks. Notably, RDMA consistently achieves the highest or comparable F1 across all configurations regardless of the underlying LLM.}

\rev{\textbf{Larger models are too expensive to deploy locally on large clinical corpuses.} Estimating mining costs on MIMIC-IV notes \cite{johnson2023mimic4_note} based on run time and GPU rental prices, we observe that larger non-quantized LLMs are substantially more expensive to run (i.e up to 10x) than their smaller quantized counterparts in Fig.~\ref{fig:performance_cost}.}

\rev{\textbf{Additional reasoning steps enable smaller quantized models to surpass larger ones.} While the performance gap between RAG-based approaches and RDMA narrows as model size increases in Fig.~\ref{fig:rd_model_perf_across_sizes}, smaller models remain substantially cheaper to run. As illustrated in Fig.~\ref{fig:performance_cost}, our quantized 24B Mistral model exceeds the performance of the much larger non-quantized Llama 3.3 70B at a fraction of the cost. Moreover, as shown in Fig.~\ref{fig:rd_model_perf_across_sizes}, RDMA performance is comparable to our cloud-based GPT-5 baseline regardless of the underlying framework, suggesting that frontier models do not necessarily confer an advantage on rare disease mining tasks.}

\rev{\subsection{Agent-Assisted Annotations Case Study}} \label{sec: improving existing annotations}
\rev{We use our cleaning of the MIMIC3-RD annotations as a case study to explore how agentic systems can contribute beyond robust entity mining—specifically, by reducing annotator burden while preserving annotation quality. We compare two correction strategies: one in which a human annotator reviews all annotations, and one in which the annotator reviews only those flagged by the agent.}

\textbf{Accelerating annotation workflows.} \rev{RDMA supports clinical experts by surfacing retrieved candidate entities, contextual information, and previously annotated Orphacodes in a streamlined format, directly addressing the time constraints faced by busy clinicians.}

\rev{To focus expert attention where it matters most, RDMA employs a two-stage verification process. First, it compares its mining results against prior annotations to compute pseudo false negatives, false positives, and true positives. Second, a verifier module flags the most contentious cases: false negatives that are not rare diseases, false positives that are rare diseases, and true positives that are not rare diseases, precisely the cases where human judgment is most valuable.}

\rev{This targeted review reduces annotation burden by 63\%, decreasing the number of cases requiring re-review from 333 to 122 (Table~\ref{tab:human_v_rdma&human_stats}), while maintaining high agreement with fully human-annotated references (Table~\ref{tab:rdma_agreement}). As a qualitative example, Appendix Fig.~\ref{fig:ex_annotation} shows RDMA correctly identifying that ``NPH'' had been mislabeled as a rare disease; the surrounding context instead pointed to insulin resistance, a correction subsequently confirmed by human experts.}

\textbf{RDMA identifies overlooked disease mentions.} \rev{Beyond reducing workload, RDMA also improves annotation completeness. As shown in Table~\ref{tab:human_v_rdma&human_stats}, RDMA-assisted annotation uncovers a greater number of unique rare diseases than human annotation alone, with at least 10 rare disease mentions present in the clinical notes going undetected during initial human review. This highlights RDMA's dual role: as a validation tool that audits existing annotations, and as a discovery mechanism that recovers entities missed in preliminary efforts. Qualitative examples of such overlooked entities are provided in Appendix Fig.~\ref{fig:annotator_agreement_qualitative} and Fig. \ref{fig:Top_Corrections}.}

\rev{Taken together, these results demonstrate that agentic systems can simultaneously reduce annotator workload and improve the accuracy and completeness of the resulting annotations. For more information, annotation guidelines are detailed in Appendix~\ref{appendix:annotator guidelines}.}

\begin{table*}[h!]
\centering
\begin{minipage}[t]{0.54\textwidth}
    \centering
    \resizebox{\textwidth}{!}{%
    \begin{tabular}{l c c c}
    \hline
    \textbf{Metric} & \textbf{Initial} & \textbf{Human Only} & \textbf{RDMA \& Human} \\
    \hline
    Total Docs & 117 & 117 & 117 \\
    \textbf{Annotations Re-reviewed} & N/A & \textbf{333} & \textbf{122} \\
    \textbf{Unique Rare Diseases} & 192 & \textbf{120} & \textbf{135} \\
    \hline
    \end{tabular}}
    \caption{Dataset comparison across annotation stages. Human correction reduces unique rare diseases from 192 to 120 ($-$37.5\%), likely removing false positives. RDMA \& Human recovers annotations to 135 ($+$12.5\% from human-only), capturing valid rare diseases missed initially. \textbf{RDMA reduces annotations requiring human re-review by over 63\%.}}
    \label{tab:human_v_rdma&human_stats}
\end{minipage}
\hfill
\begin{minipage}[t]{0.42\textwidth}
    \centering
    \resizebox{\textwidth}{!}{%
    \begin{tabular}{lcc}
    \toprule
    \textbf{Metric} & \textbf{RDMA Only} & \textbf{RDMA \& Human} \\
    \midrule
    Cohen's Kappa & 0.46 & 0.81 \\
    F1 Score      & 0.74 & 0.94 \\
    Precision     & 0.92 & 0.92 \\
    Recall        & 0.62 & 0.96 \\
    \bottomrule
    \end{tabular}}
    \caption{\rev{\textbf{Human in the Loop Refinement via Two Stage Discrepancy Audit.} In step 4 of RDMA, the agent acts as a tie breaker analyzing entity sets from two separate agentic passes, computing pseudo FP, FN, and TP labels before auditing its own classifications to flag high-uncertainty cases for human review. After final human review, agreement with expert annotators remains high, suggesting that the flagging accurately filters for the most important entities needing re-review.}}
    \label{tab:rdma_agreement}
\end{minipage}
\end{table*}

\section{Discussion} \label{sec: discussion}
\rev{The RDMA framework addresses critical bottlenecks in clinical information extraction by leveraging the adaptability of agentic workflows. This section explores the strengths of our approach, the persistent challenges of medical reasoning, and the need for more robust benchmarking in rare disease research.}

\begin{wraptable}{r}{0.45\textwidth}
    \centering
    \small 
    \vspace{-15pt} 
    \begin{tabular}{lccc}
        \toprule
        \textbf{\rev{Baseline}} & \textbf{\rev{F1}} & \textbf{\rev{P}} & \textbf{\rev{R}} \\
        \midrule
        \rev{RDMA} & \rev{\textbf{0.65}} & \rev{0.63} & \rev{0.68} \\
        \rev{No Lab Tool} & \rev{0.64} & \rev{0.63} & \rev{0.64} \\
        \rev{No Impl. Check} & \rev{0.65} & \rev{\textbf{0.64}} & \rev{0.65} \\
        \rev{No Verif.} & \rev{0.53} & \rev{0.42} & \rev{\textbf{0.71}} \\
        \bottomrule
    \end{tabular}
    \caption{\rev{\textbf{RDMA Ablation Study.} Impact of individual components on phenotype mining performance of a quantized Mistral 24B model on the CSC \cite{garcia2024_HPORAG}.}}
    \label{tab:rdma_ablation}
    \vspace{-20pt} 
\end{wraptable}

\rev{\textbf{The flexibility of agentic frameworks.} A key advantage of agentic frameworks over fixed pipelines is their task-specific adaptability. As shown in Fig. \ref{fig:RDMAvRAG}, phenotype and disease extraction agents utilize distinct tools while maintaining a shared workflow of extract, verify, match, and refine. RDMA allows for modular customization—such as including abbreviation detection or lab event databases for phenotype extraction while omitting lab events for rare disease extraction, where they provide no value. Our ablation study (Table \ref{tab:rdma_ablation}) underscores the importance of this modularity. While the lab test tool and implication reasoning provide marginal gains in recall, the \textbf{verification step} is the most critical component, providing a substantial boost to precision by filtering out hallucinations and misalignment.}

\begin{figure}[h!]
    \centering
    \includegraphics[width=1.0\textwidth]{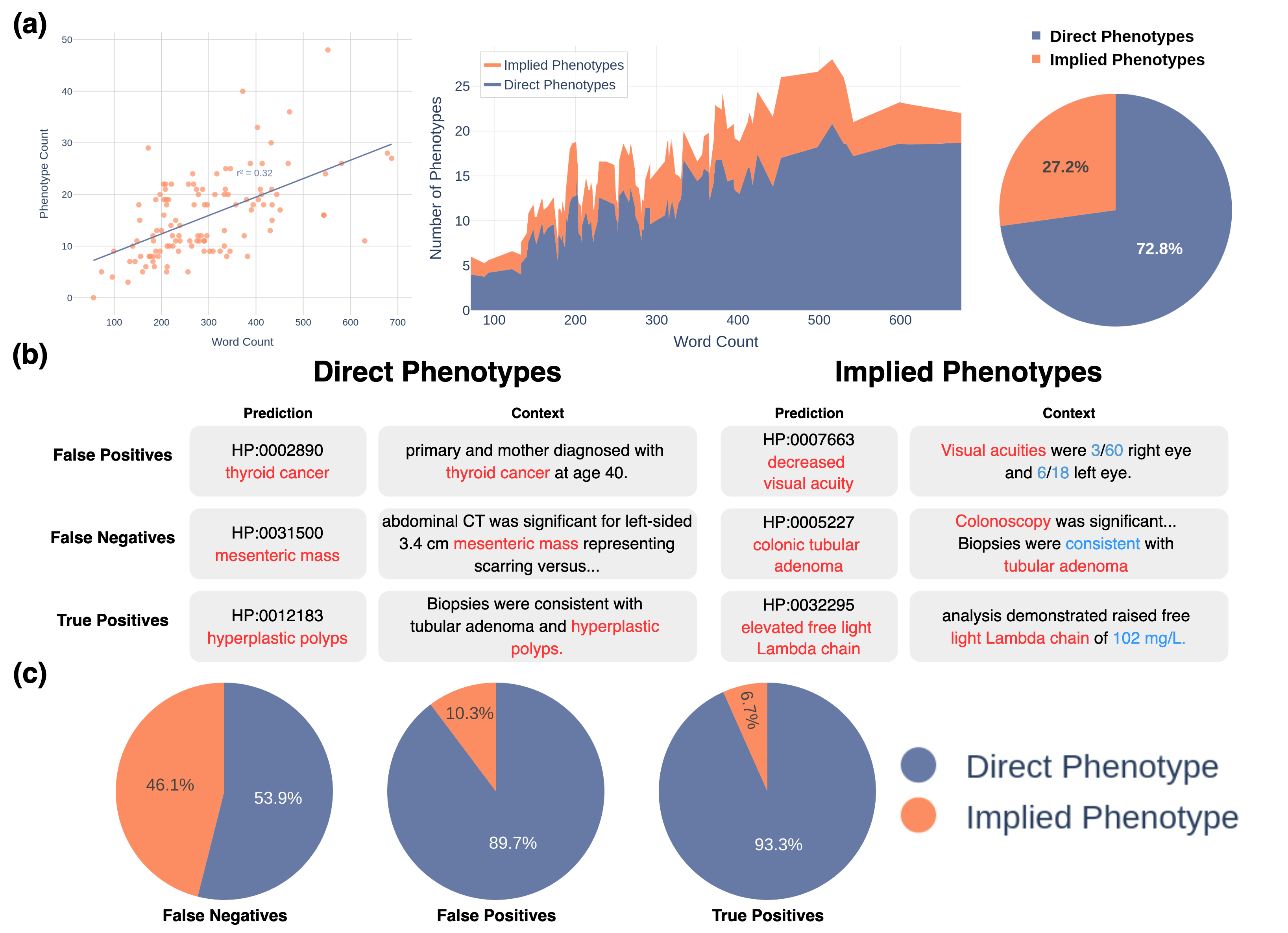}
    \caption{\rev{\textbf{CSC Phenotype Extraction Qualitative Analysis.} We evaluate on 116 clinical cases (33,824 words total) from \cite{garcia2024_HPORAG}. \textbf{(a) Dataset Statistics.} Document length weakly correlates with phenotype count. Of 1,813 total phenotypes, 1,320 (72.8\%) appear explicitly while 493 (27.2\%) are implied. \textbf{(b) Qualitative Analysis.} Our method captures phenotypes implied by lab tests. Key challenges include: (1) False positives that are clinically valid but lack exact HPO matches; (2) False negatives arising from missing non-obvious entities or failing to capture long-range context. RDMA also identifies direct phenotypes like ``thyroid cancer'' that human annotators missed. \textbf{(c) Performance Breakdown.} Implied phenotypes comprise only 27.2\% of labels but account for a disproportionate share of false negatives.}}
    \label{fig:DatasetBreakdown}
\end{figure}

\rev{\textbf{Challenges in phenotype mining and medical reasoning.} Despite the strengths of agentic systems, significant challenges remain in capturing \textbf{implied phenotypes}. Analysis of our benchmark (Fig.~\ref{fig:DatasetBreakdown}) reveals that over 25\% of phenotypes are not explicitly mentioned in the text. These require LLMs to infer conditions from lab results or symptom clusters. Current LLMs remain insufficient for basic reasoning tasks like implying phenotypes related to lab events, as shown by the tiny improvements in Fig.~\ref{fig:DatasetBreakdown}. RDMA still struggles with: (1) \textit{Long-context reasoning}: failing to link distal mentions (e.g., a colonoscopy performed earlier in a note to a ``colonic tubular adenoma''); (2) \textit{Terminology nuance}: identifying ``decreased visual acuity'' as a match for the HPO code ``reduced visual acuity''; and (3) \textit{Negation detection}: occasionally ignoring negative qualifiers (e.g., ``negative for...''), even when prompted to consider negations.}

\begin{figure}[h!]
    \centering
    \includegraphics[width=\textwidth]{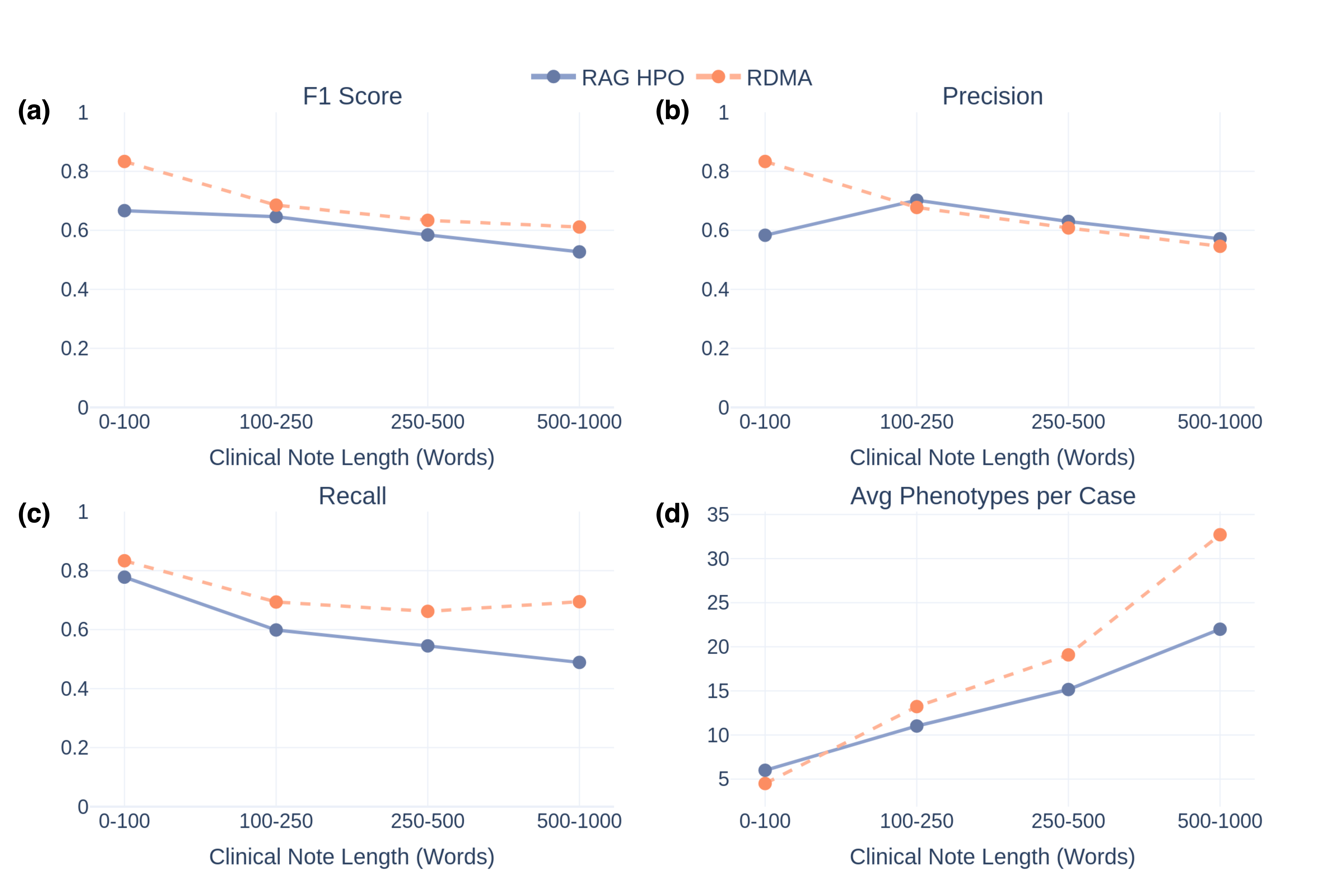}
    \caption{\rev{\textbf{RDMA outperforms local RAG-HPO across note lengths.} Real-world clinical notes (MIMIC-IV) are substantially longer than annotated case studies (median 1,320 words vs 271.5 words). While overall precision slightly decreases as note length increases (b), RDMA maintains stable recall performance (c), enabling it to outperform its RAG-based counterpart across all note lengths (a). Note count (d) shows phenotype count typically increases with document length.}}
    \label{fig:RDMA_Note_Lengths}
\end{figure}

\rev{\textbf{Robustness to clinical note complexity.} A primary hurdle in rare disease mining is the discrepancy between scientific passages and real-world EHRs. Clinical notes in datasets like MIMIC-IV are often thousands of words long and laden with noise such as abbreviations and misspellings. As shown in Fig. \ref{fig:RDMA_Note_Lengths}, RDMA demonstrates remarkable robustness to document length. Unlike one-shot RAG approaches that typically perform a single extraction and degrade as context grows, RDMA’s recall slightly increases with longer texts. By retrieving against individual sentences and employing agentic verification, we maintain a stable F1 score across varying lengths. Nonetheless, a great area of interest is improving the ability of these systems at longer and noisier contexts, especially given that the performance discrepancy of approaches on the scientifc RareDis corpus and our cleaned MIMIC3 clinical note corpus has a gap of over 20\% F1 as shown in Table \ref{tab:main_results_combined_v3}.  }

\rev{\textbf{The need for comprehensive rare disease benchmarks.} Current datasets for mining \cite{lobo2017identifying_human_phenotype_terms,garcia2024_HPORAG,martínezdemiguel2021rarediscorpuscorpusannotated} and differential diagnosis (DDx) \cite{chen2024rarebenchllmsserverare,chen2025rareagentsadvancingraredisease, germain2025applyingAI_RD} face significant limitations. Sentence snippets and scientific passages may not reflect the lengthy, noisy nature of EHRs. Furthermore, many DDx benchmarks use ICD codes as ground-truth proxies, which are flawed because they do not capture rare diseases precisely enough and are often missing from the text \cite{mazzucato2023orphacodes_better, cheng-etal-2023-mdace}. Expecting LLMs to perform accurate DDx without the ability to extract all phenotypes reliably is premature. RDMA provides a framework to generate more accurate ORPHA and HPO-based profiles, but the field requires larger, expert-verified datasets that encompass the full diversity of the Orphanet ontology \cite{weinreich2008orphanet} and HPO \cite{gargano2024mode_hpo}. For future work, extending our mining approach with human verification on the entire MIMIC-III \cite{johnson2016mimic3} and MIMIC-IV \cite{johnson2023mimic4} corpus would be ideal. Gold-standard rare disease and phenotype annotations do not currently exist for MIMIC-IV at scale, however, making this a critical next step. }
\section{Methodology} \label{sec:methodology}

\rev{\textbf{Problem Formulation.} RDMA operates over a corpus of clinical documents and aims to identify two types of entities within each document: phenotypes, represented as HPO codes from the Human Phenotype Ontology \cite{gargano2024mode_hpo}, and rare diseases, represented as ORPHA codes from the Orphanet ontology \cite{weinreich2008orphanet}. The output is an annotated dataset where each document is paired with the set of phenotypes and rare diseases found within it.}

\rev{\textbf{Tool Retrieval.} Most of RDMA's tools are structured as retrieval systems over medical databases. Given a candidate entity string, we embed it and retrieve the most semantically similar entries from the relevant ontology database using Euclidean distance over embeddings. We use FAISS \cite{douze2025faisslibrary} for efficient search and MedEmbed Small \cite{balachandran2024medembed} as our embedding model, chosen for its strong performance on medical text.}

\rev{\textbf{RDMA Overview.} Algorithm \ref{alg:rdma} presents the full RDMA pipeline. At a high level, RDMA processes each clinical document sentence by sentence, using semantic search to retrieve ontology candidates and an LLM to extract entities of interest. Each candidate entity then passes through a multi-step verification pipeline before being mapped to a formal ontological code. Finally, the resulting annotations are compared against a prior annotation set, with disagreements flagged for human expert review. Not all steps apply to every entity type, phenotype extraction involves lab event detection and implied phenotype generation, while rare disease extraction focuses on abbreviation expansion and confirming that matched entries are true diseases rather than related biological entities. We describe each step below.}

\begin{algorithm}[H]
\caption{RDMA: Mining Pipeline}
\label{alg:rdma}
{\small
\begin{algorithmic}[1]
\Require Corpus $X$, ontologies $\mathcal{O} = \mathcal{D}_{HPO} \cup \mathcal{D}_{Orphanet}$, mode $m \in \{\text{rare}, \text{phenotype}, \text{both}\}$
\Ensure Annotated dataset $D$
\For{each document $x \in X$}
    \For{each sentence $s$ in $x$}
        \State $cands_s \gets \Call{SemanticSearch}{s,\ \mathcal{O}}$
        \State $E \gets \Call{LLMExtract}{s,\ cands_s}$
    \EndFor
    \For{each entity $e \in E$}
        \If{$m \in \{\text{rare}, \text{both}\}$}
            \State $e \gets \Call{ExpandAbbreviation}{e}$
            \If{$\Call{OntologyMatch}{e,\ s}$ \textbf{ and } $\Call{LLMIsDisease}{e}$}
                \State Accept $e$ as verified rare disease
            \EndIf
        \EndIf
        \If{$m \in \{\text{phenotype}, \text{both}\}$}
            \If{$\Call{OntologyMatch}{e,\ s}$}
                \State Accept $e$ as verified phenotype
            \ElsIf{$\Call{IsAbnormalLabValue}{e}$}
                \State Accept $\Call{LLMGenerateDirection}{e}$ as implied phenotype
            \ElsIf{$\Call{LLMImpliesPhenotype}{e}$ is valid in $\mathcal{D}_{HPO}$}
                \State Accept implied phenotype
            \EndIf
        \EndIf
    \EndFor
    \For{each verified entity $e$}
        \State $cands_e \gets \Call{SemanticSearch}{e,\ \mathcal{O}}$
        \State $code \gets \Call{LLMMatch}{e,\ s,\ cands_e}$
    \EndFor
\EndFor
\State \Return $D$
\end{algorithmic}
}
\end{algorithm}

\begin{algorithm}[H]
\caption{RDMA: Dataset Refinement}
\label{alg:rdma_refinement}
{\small
\begin{algorithmic}[1]
\Require $D$, prior annotations $\mathcal{A}_{prior}$
\Ensure Flagged entities $\mathcal{F}$ for review
\State $\mathcal{A}_{prior} \gets \Call{KeywordFilter}{\mathcal{A}_{prior}}$
\State $TP, FN, FP \gets \Call{Compare}{D,\ \mathcal{A}_{prior}}$
\For{each $e \in TP \cup FN \cup FP$}
    \State $v \gets \Call{OntologyMatch}{e}$
    \If{($e \in TP$ \textbf{ and not } $v$) \textbf{ or }}
    \StatexX{\ \ \ ($e \in FN$ \textbf{ and } $v$) \textbf{ or }}
    \StatexX{\ \ \ ($e \in FP$ \textbf{ and } $v$)}
        \State $\mathcal{F} \gets \mathcal{F} \cup \{e\}$
    \EndIf
\EndFor
\State \Return $\mathcal{F}$
\end{algorithmic}
}
\end{algorithm}

\subsection{Entity Extraction}
\rev{Each document is split into sentences, and for each sentence we query the relevant ontology database to retrieve the top-5 most similar phenotype or disease candidates. Both the sentence and its candidates are passed to an LLM via $\Call{LLMExtract}{}$, which identifies any phenotype or rare disease entities worth investigating further. Sentence-level chunking balances specificity against computational cost: finer chunks yield more precise retrieval but increase processing time, a trade-off we examine further in Appendix \ref{appendix: sentence agglomeration}.}

\subsection{Entity Verification and Implication} \label{sec:verify-and-implication-reasoning}
\rev{Each extracted entity passes through a verification pipeline before being accepted. The steps differ for phenotypes versus rare diseases (see general structure in Algorithm \ref{alg:rdma}, lines 7--20).}

\textbf{Abbreviation expansion.} \rev{For rare disease candidates, we first attempt to expand any clinical abbreviations via $\Call{ExpandAbbreviation}{}$ before proceeding, since abbreviation ambiguity is particularly problematic in this setting.}

\textbf{Direct ontology verification.} \rev{We call $\Call{OntologyMatch}{}$ to check whether the entity genuinely corresponds to an HPO or Orphanet entry in context. For rare diseases, we additionally call $\Call{LLMIsDisease}{}$ to confirm the matched entry is a disease rather than a related entity type (e.g., a gene, protein, or enzyme) that also appears in Orphanet.}

\textbf{Lab event detection and implication (phenotypes only).} \rev{If a phenotype entity is not directly verified, we check whether it describes an abnormal lab measurement via $\Call{IsAbnormalLabValue}{}$. If so, $\Call{LLMGenerateDirection}{}$ converts it into the corresponding implied phenotype (e.g., elevated creatinine $\to$ hypercreatininemia).}

\textbf{Implied phenotype generation (phenotypes only).} \rev{For entities that are neither direct phenotypes nor lab events, $\Call{LLMImpliesPhenotype}{}$ checks whether the entity contextually implies a phenotype. If so, we generate and verify the implied phenotype against the HPO ontology before accepting it.}

\subsection{Verified Entity Matching}
\rev{Each verified entity is mapped to a specific ontological code by $\Call{LLMMatch}{}$, which is presented with the entity, its source sentence, and the top retrieved ontology candidates. This follows a similar workflow to RAG-HPO \cite{garcia2024_HPORAG}, extended here to cover rare disease identification.}

\subsection{Dataset Refinement} \label{sec: methodology - dataset refinement}
\rev{After assembling mined annotations, RDMA compares them against a prior annotation set using Algorithm \ref{alg:rdma_refinement}. Entities that agree across both sources are treated as likely correct; disagreements are flagged via $\Call{FlagForReview}{}$ for human expert judgment. The flagging logic is asymmetric by design: a true positive that fails re-verification, a false negative that passes, or a false positive that passes are all suspicious and warrant review, while cases where both sources consistently agree or disagree are left unflagged.}

\rev{\textbf{Preliminary Filtering.} Before running RDMA refinement, we apply a set of rule-based corrections via $\Call{KeywordFilter}{}$, informed by a clinician review of the prior annotations (Appendix Table \ref{tab:Filtering_Details_ex}). This removed over 700 misannotated entries --- for example, ``MR'' had been tagged as ``Multicentric reticulohistiocytosis'' when it almost universally refers to magnetic resonance in context --- reducing the dataset from 1,073 annotations across 312 notes to 333 annotations across 117 notes (Appendix Table \ref{tab:filtering_results_ex}). For more information on the methodology such as prompts and our semantic retrieval formulation, see Appendices \ref{appendix: exp method} and \ref{appendix:prompts}.}

\backmatter
\bmhead{Acknowledgements}
This study was funded by Jump ARCHES endowment awarded by the Healthcare Engineering Systems Center at the University of Illinois Urbana-Champaign (UIUC) and Order of Saint Francis (OSF) Foundation. The funder played no role in study design, data collection, analysis and interpretation of data, or the writing of this manuscript. 

\section*{Declarations}
\begin{itemize}
\item Competing interests: All authors declare no financial or non-financial competing interests. 
\item Data availability: The datasets generated and/or analysed during the current study are available in the RDMA repository at \url{https://github.com/jhnwu3/RDMA}.
\item Code availability: The underlying code for this study is available in the RDMA repository and can be accessed via this link \url{https://github.com/jhnwu3/RDMA}.
\item Author contributions: JW wrote the manuscript, designed the figures, and implemented the experiments. AC and JS participated in the design, analysis, and discussion of the experiments. AC and JS also assisted with revisions to the manuscript. All authors read and approved the final manuscript.
\end{itemize}











\noindent






\begin{appendices}

\section{Inference and Hardware Cost Estimates} \label{appendix: cost calculations}
\rev{Here, we describe our hardware and inference costs within the context of our benchmark statistics.} While local hardware costs are based directly on the listing available for commonplace workstations, we follow a simple equation to compute our cost metrics in benchmarking phenotype extraction in Tables \ref{tab:gpu_costs} and \ref{tab:cost_calculations}. We discuss local hardware costs to run these systems in Table \ref{tab:cost_categories}. \rev{For cost reasons, our downstream performance benchmarking was done with the 4-bit quantized variants of each model when the model sizes are greater than 8B parameters. We only benchmark costs for non-quantized variants for the Llama 3.3 70B as our largest locally deployable model. Quantization was done with the bitsandbytes package \cite{dettmers2023qlora_bits}. We benchmark primarily on our case study cohort (CSC) benchmark.}

\begin{table}[h]
\centering
\begin{tabular}{|l|c|}
\hline
\textbf{GPU} & \textbf{Rental Cost (\$/hr)} \\
\hline
A6000 & 0.5 \\
\hline
RTX 3090 & 0.1 \\
\hline
\end{tabular}
\caption{GPU rental costs per hour.}
\label{tab:gpu_costs}
\end{table}

\begin{table}[ht]
\centering
\begin{tabular}{lcr}
\hline
\textbf{Category} & \textbf{GPU Configuration} & \textbf{Estimated Cost} \\
\hline
Very Low & N/A & \$120 \\
Low & 1$\times$3090 & \$2,200 \\
Medium & 1$\times$A6000 & \$6,520 \\
High & 4$\times$A6000 & \$38,500 \\
\hline
\end{tabular}
\caption{\textbf{Local Hardware Cost Categories.} These cost categories are referenced in the performance comparison in our discussion. We note that the approximate costs are based from current workstation prices \cite{newegg_3d5} \cite{newegg_velztorm} \cite{thinkmate_gpx} as of April 26, 2025. In principle, rule-based approaches do not need a GPU to run scalably where even a Raspberry Pi is computationally sufficient.}
\label{tab:cost_categories}
\end{table}

\begin{table}[h]
\centering
\begin{tabular}{|l|l|c|}
\hline
\textbf{Variable Name} & \textbf{Statistic} & \textbf{Total} \\
\hline
TN & Total Notes & 331,794 \\
\hline
MNL & Median MIMIC-IV Note Length (word count) & 1,320 \\
\hline
MCSL & Benchmark Median Case Study Length (word count) & 271.5 \\
\hline
\end{tabular}
\caption{MIMIC4 Notes Statistics.}
\label{tab:mimic4_stats}
\end{table}

\begin{table}[h]
\centering
\renewcommand{\arraystretch}{2.4}
\begin{tabular}{|l|l|c|>{\centering\arraybackslash}p{4cm}|}
\hline
\textbf{Baseline} & \textbf{GPU} & \textbf{Run Time (m)} & \textbf{Cost Calculation} \\
\hline
RAG-HPO (Mistral 24B$^q$) & 1$\times$RTX 3090 & 39 & $\displaystyle\frac{39 \times 0.1}{60} \times \frac{\text{MNL} \times \text{TN}}{\text{MCSL}}$ \\
\hline
RAG-HPO (Llama 3.3-70B$^q$) & 1$\times$A6000 & 62 & $\displaystyle\frac{62 \times 0.5}{60} \times \frac{\text{MNL} \times \text{TN}}{\text{MCSL}}$ \\
\hline
RAG-HPO (Llama 3.3-70B) & 4$\times$A6000 & 70 & $\displaystyle\frac{70 \times 4 \times 0.5}{60} \times \frac{\text{MNL} \times \text{TN}}{\text{MCSL}}$ \\
\hline
RDMA & 1$\times$RTX 3090 & 121 & $\displaystyle\frac{121 \times 0.1}{60} \times \frac{\text{MNL} \times \text{TN}}{\text{MCSL}}$ \\
\hline
\end{tabular}
\caption{Baseline comparison of different methods with their runtime and cost calculations.}
\label{tab:cost_calculations}
\end{table}

\begin{table}[h]
\centering
\caption{\rev{Runtime comparison on CSC dataset using Mistral 24B 4-bit quantized. Given that the CSC dataset contains 32,260 words, we have a pipeline that can extract, verify, and match HPO entiies at a rate of approximately 278 words a minute on an RTX 3090.}}
\label{tab:runtime_mistral24b_rdma_raghpo}
\begin{tabular}{lccc}
\toprule
\textbf{Method} & \textbf{Extraction (m)} & \textbf{Verification (m)} & \textbf{Matching (m)} \\
\midrule
RAG HPO & 15 & --- & 24 \\
RDMA    & 68 & 29  & 24 \\
\bottomrule
\end{tabular}
\end{table}


\begin{table}[t]
\centering
\caption{\rev{Statistics of rare disease NER benchmarks and phenotype datasets used for evaluation. RareDis splits: train (711 docs, 3{,}608 entities, 111{,}369 words), dev (97 docs, 525 entities, 15{,}268 words), test (203 docs, 1{,}088 entities, 31{,}042 words). For BiolarkGSC, oru phenotype training corpus, we use a random 80-10-10 split, roughly resulting in 182 train / ~23 dev / ~23 test documents. For all general LLM-based baselines, we use the entire corpus for benchmarking as they do not require training. }}
\label{tab:datasets}
\begin{tabular}{llrrrrr}
\toprule
\textbf{Category} & \textbf{Dataset} & \textbf{Docs} & \textbf{Entities} & \textbf{Total Words} & \textbf{Avg.\ Words} & \textbf{Max Words} \\
\midrule
\multirow{2}{*}{Phenotype} 
    & BioLarkGSC \cite{yang2023enhancingphenotyperecognitionclinical} & 228     & 2{,}773 &  33{,}942  & 148.9   & 359   \\
    & CSC  \cite{garcia2024_HPORAG}      & 116     & 1{,}813 &  32{,}260  & 278.1   & 675   \\
\midrule
\multirow{2}{*}{Rare Disease}
    & RareDis  \cite{martinez2022raredis}  & 1{,}011 & 5{,}221 & 157{,}679  & 156.0   & 601   \\
    & MIMIC3-RD \cite{johnson2016mimic3} & 117     &   176   & 221{,}980  & 1{,}897.3 & 6{,}726 \\
\bottomrule
\end{tabular}
\end{table}

\clearpage

\section{Performance Metric Calculations} \label{appendix: Performance Metric Calculations}
For each clinical document $x_i$, we compare the set of ground-truth human-annotated codes with the set of predicted codes from our RDMA framework. Let $\Phi_P(x_i)$ and $\Phi_R(x_i)$ denote the ground-truth phenotype and rare disease codes for document $x_i$, respectively, and let $\hat{\Phi}_P(x_i)$ and $\hat{\Phi}_R(x_i)$ denote the corresponding predicted codes.

For each document $x_i$ and each task (phenotypes or rare diseases), we define:

\textbf{True Positives:}
\begin{align}
TP_P^i &= |\hat{\Phi}_P(x_i) \cap \Phi_P(x_i)| \\
TP_R^i &= |\hat{\Phi}_R(x_i) \cap \Phi_R(x_i)|
\end{align}

\textbf{False Positives:}
\begin{align}
FP_P^i &= |\hat{\Phi}_P(x_i) \setminus \Phi_P(x_i)| \\
FP_R^i &= |\hat{\Phi}_R(x_i) \setminus \Phi_R(x_i)|
\end{align}

\textbf{False Negatives:}
\begin{align}
FN_P^i &= |\Phi_P(x_i) \setminus \hat{\Phi}_P(x_i)| \\
FN_R^i &= |\Phi_R(x_i) \setminus \hat{\Phi}_R(x_i)|
\end{align}

We then aggregate these counts across all $n$ documents in our corpus to compute overall performance metrics:

\begin{align}
TP_P &= \sum_{i=1}^{n} TP_P^i, \quad FP_P = \sum_{i=1}^{n} FP_P^i, \quad FN_P = \sum_{i=1}^{n} FN_P^i \\
TP_R &= \sum_{i=1}^{n} TP_R^i, \quad FP_R = \sum_{i=1}^{n} FP_R^i, \quad FN_R = \sum_{i=1}^{n} FN_R^i
\end{align}

For phenotype extraction:
\begin{align}
\text{Precision}_P &= \frac{TP_P}{TP_P + FP_P} \\
\text{Recall}_P &= \frac{TP_P}{TP_P + FN_P} \\
\text{F1}_P &= 2 \cdot \frac{\text{Precision}_P \cdot \text{Recall}_P}{\text{Precision}_P + \text{Recall}_P}
\end{align}

For rare disease extraction:
\begin{align}
\text{Precision}_R &= \frac{TP_R}{TP_R + FP_R} \\
\text{Recall}_R &= \frac{TP_R}{TP_R + FN_R} \\
\text{F1}_R &= 2 \cdot \frac{\text{Precision}_R \cdot \text{Recall}_R}{\text{Precision}_R + \text{Recall}_R}
\end{align}


\section{Bootstrapped Confidence Intervals} \label{appdx: CIs}
Because reruns are prohibitively expensive, we minimize LLM temperature (i.e t=0.001) to limit variance. However, we still include bootstrapped confidence intervals (Tables \ref{tab:phenotype_ci} and \ref{tab:raredis_ci}).


\begin{table*}[t]
\centering\small\setlength{\tabcolsep}{5pt}
\caption{Phenotype extraction results. F1$^{\pm}$ denotes the half-width of the 95\% bootstrap CI.
BioLARK-GSC: $n{=}228$ docs; CSC: $n{=}116$ docs.
Zeroshot LLMs score 0 on both datasets (cannot generate HPO identifiers without structured prompting).
Best per approach group \textbf{bolded}.}
\label{tab:phenotype_ci}
\begin{tabular}{llcc}
\toprule
\textbf{Approach} & \textbf{Model} & \textbf{BioLarkGSC+} & \textbf{CSC} \\
\midrule
\multicolumn{4}{l}{\textit{Baselines}} \\
& BioBERT-MRC & 0.498$^{\pm0.064}$ & 0.382$^{\pm0.030}$ \\
& BioClinicalBERT & 0.441$^{\pm0.081}$ & 0.392$^{\pm0.031}$ \\
& FastHPOCR & 0.603$^{\pm0.023}$ & 0.480$^{\pm0.029}$ \\
& Dictionary & 0.326$^{\pm0.025}$ & 0.319$^{\pm0.025}$ \\
& i2b2 & 0.491$^{\pm0.024}$ & 0.461$^{\pm0.031}$ \\
& PhenoGPT & 0.427$^{\pm0.027}$ & 0.373$^{\pm0.035}$ \\
\midrule
\multicolumn{4}{l}{\textit{Zeroshot}} \\
& GPT-5 & 0.000$^{\pm.000}$ & 0.000$^{\pm.000}$ \\
& Llama-3 70B & 0.000$^{\pm.000}$ & 0.000$^{\pm.000}$ \\
& Llama-3 8B & 0.000$^{\pm.000}$ & 0.000$^{\pm.000}$ \\
& Mistral 24B & 0.000$^{\pm.000}$ & 0.000$^{\pm.000}$ \\
& Qwen3 32B & 0.000$^{\pm.000}$ & 0.000$^{\pm.000}$ \\
\midrule
\multicolumn{4}{l}{\textit{RAG}} \\
& GPT-5 & \textbf{0.525}$^{\pm0.027}$ & 0.543$^{\pm0.027}$ \\
& Llama-3 70B & 0.474$^{\pm0.026}$ & \textbf{0.624}$^{\pm0.025}$ \\
& Llama-3 8B & 0.345$^{\pm0.024}$ & 0.509$^{\pm0.033}$ \\
& Mistral 24B & 0.398$^{\pm0.026}$ & 0.595$^{\pm0.034}$ \\
& Qwen3 32B & 0.401$^{\pm0.024}$ & 0.594$^{\pm0.030}$ \\
\midrule
\multicolumn{4}{l}{\textit{RDMA}} \\
& GPT-5 & 0.545$^{\pm0.029}$ & 0.646$^{\pm0.041}$ \\
& Llama-3 70B & 0.484$^{\pm0.023}$ & 0.627$^{\pm0.027}$ \\
& Llama-3 8B & 0.331$^{\pm0.019}$ & 0.519$^{\pm0.033}$ \\
& Mistral 24B & \textbf{0.559}$^{\pm0.021}$ & \textbf{0.657}$^{\pm0.029}$ \\
& Qwen3 32B & 0.473$^{\pm0.022}$ & 0.649$^{\pm0.028}$ \\
\bottomrule
\end{tabular}
\end{table*}

\begin{table*}[t]
\centering\small\setlength{\tabcolsep}{4pt}
\caption{Rare disease mention extraction results. F1$^{\pm}$ denotes the half-width of the 95\% bootstrap CI.
MIMIC3-RD Entity: $n{=}117$ docs; MIMIC3-RD Code: $n{=}79$ docs; RareDis: $n{=}1{,}011$ docs.
Best per approach group \textbf{bolded}.}
\label{tab:raredis_ci}
\begin{tabular}{llccc}
\toprule
\textbf{Approach} & \textbf{Model} & \textbf{MIMIC-III (text)} & \textbf{MIMIC-III (code)} & \textbf{RareDis} \\
\midrule
\multicolumn{5}{l}{\textit{Baselines}} \\
& BioBERT-MRC & 0.046$^{\pm0.011}$ & 0.046$^{\pm0.012}$ & 0.717$^{\pm0.027}$ \\
& BioClinicalBERT & 0.050$^{\pm0.011}$ & 0.050$^{\pm0.013}$ & 0.673$^{\pm0.032}$ \\
& Dictionary & 0.431$^{\pm0.074}$ & 0.384$^{\pm0.066}$ & 0.555$^{\pm0.022}$ \\
& i2b2 & 0.020$^{\pm0.004}$ & 0.019$^{\pm0.004}$ & 0.260$^{\pm0.012}$ \\
\midrule
\multicolumn{5}{l}{\textit{Zeroshot}} \\
& GPT-5 & \textbf{0.497}$^{\pm0.069}$ & \textbf{0.098}$^{\pm0.056}$ & \textbf{0.810}$^{\pm0.016}$ \\
& Llama-3 70B & 0.223$^{\pm0.038}$ & 0.000$^{\pm.000}$ & 0.705$^{\pm0.019}$ \\
& Llama-3 8B & 0.134$^{\pm0.024}$ & 0.000$^{\pm.000}$ & 0.548$^{\pm0.020}$ \\
& Mistral 24B & 0.228$^{\pm0.041}$ & 0.002$^{\pm0.004}$ & 0.704$^{\pm0.020}$ \\
& Qwen3 32B & 0.326$^{\pm0.066}$ & 0.000$^{\pm.000}$ & 0.651$^{\pm0.022}$ \\
\midrule
\multicolumn{5}{l}{\textit{RAG}} \\
& GPT-5 & 0.505$^{\pm0.069}$ & \textbf{0.391}$^{\pm0.068}$ & \textbf{0.826}$^{\pm0.017}$ \\
& Llama-3 70B & 0.125$^{\pm0.023}$ & 0.139$^{\pm0.024}$ & 0.483$^{\pm0.018}$ \\
& Llama-3 8B & 0.047$^{\pm0.012}$ & 0.025$^{\pm0.007}$ & 0.267$^{\pm0.015}$ \\
& Mistral 24B & \textbf{0.592}$^{\pm0.068}$ & 0.032$^{\pm0.010}$ & 0.226$^{\pm0.012}$ \\
& Qwen3 32B & 0.133$^{\pm0.027}$ & 0.143$^{\pm0.030}$ & 0.748$^{\pm0.019}$ \\
\midrule
\multicolumn{5}{l}{\textit{RDMA}} \\
& GPT-5 & 0.513$^{\pm0.072}$ & 0.465$^{\pm0.083}$ & 0.780$^{\pm0.016}$ \\
& Llama-3 70B & 0.442$^{\pm0.056}$ & 0.439$^{\pm0.069}$ & \textbf{0.845}$^{\pm0.013}$ \\
& Llama-3 8B & 0.072$^{\pm0.013}$ & 0.097$^{\pm0.020}$ & 0.595$^{\pm0.016}$ \\
& Mistral 24B & \textbf{0.592}$^{\pm0.068}$ & \textbf{0.526}$^{\pm0.080}$ & 0.814$^{\pm0.014}$ \\
& Qwen3 32B & 0.528$^{\pm0.057}$ & 0.467$^{\pm0.071}$ & 0.828$^{\pm0.013}$ \\
\bottomrule
\end{tabular}
\end{table*}

To estimate the stability of micro-F1 scores without rerunning experiments,
we apply document-level bootstrap resampling over per-document evaluation
counts $(tp_i, fp_i, fn_i)$ drawn from each evaluated run.

\paragraph{Point Estimate.}
Micro-F1 is computed from aggregate counts over all $N$ documents:
\begin{align}
    \mathrm{TP} &= \sum_{i=1}^{N} tp_i, \quad
    \mathrm{FP} = \sum_{i=1}^{N} fp_i, \quad
    \mathrm{FN} = \sum_{i=1}^{N} fn_i, \notag \\
    P &= \frac{\mathrm{TP}}{\mathrm{TP} + \mathrm{FP}}, \quad
    R  = \frac{\mathrm{TP}}{\mathrm{TP} + \mathrm{FN}}, \quad
    F_1 = \frac{2PR}{P + R},
\end{align}
with each term defined as zero when its denominator is zero.

\paragraph{Resampling Procedure.}
We draw $B = 1{,}000$ bootstrap replicates. In each replicate, $N$ documents
are sampled with replacement from the original $N$; micro-F1 is recomputed
from the resampled aggregate counts, yielding a distribution
$\{F_1^{(b)}\}_{b=1}^{B}$.

\paragraph{Confidence Interval.}
A $(1-\alpha)$ confidence interval is read from the sorted bootstrap
distribution. For a 95\% CI ($\alpha = 0.05$):
\begin{equation}
    \hat{F}_1^{\,\mathrm{lo}} = F_1^{(\lfloor \alpha/2 \cdot B \rfloor)},
    \qquad
    \hat{F}_1^{\,\mathrm{hi}} = F_1^{(\lfloor (1-\alpha/2) \cdot B \rfloor)},
\end{equation}
corresponding approximately to the 2.5th and 97.5th percentiles.
All runs with fewer than 20 documents are excluded to avoid partial or
debug evaluations.

\paragraph{Rationale.}
Resampling is performed at the document level rather than the entity or token
level, as the clinical note or disease case is the natural unit of variance:
some documents are substantially harder than others, and token-level
resampling would underestimate uncertainty by treating all mentions as
independent.
\section{\rev{AI Assisted Corrections and Annotations on MIMIC-III Corpus}}
\begin{figure}[h]
    \centering
    \includegraphics[width=0.55\textwidth]{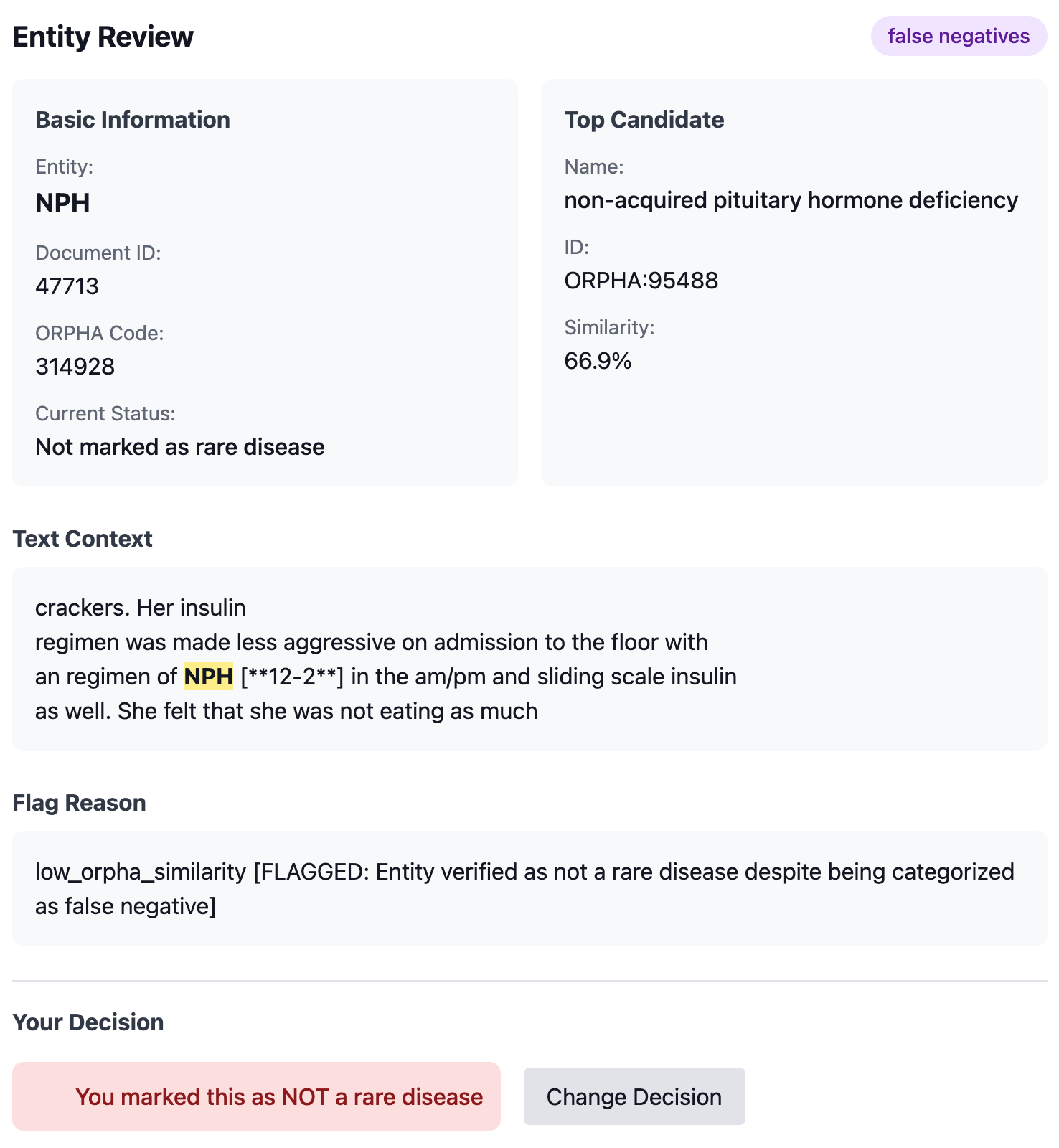}
    \caption{\textbf{Example of Inappropriate Annotation in Noisy Dataset.} We note that while NPH as an abbreivation can be related to ``normal pressure hydrocephalus" or other related conditions in the Orphanet ontology as annotated by \cite{dong2023ontology_poor_annotations}, NPH here in this context is actually referring to neutral protamine hagedorn, a type of insulin used to treat diabetes. }
    \label{fig:ex_annotation}
\end{figure}

\begin{figure}[h]
    \centering
    \includegraphics[width=0.8\textwidth]{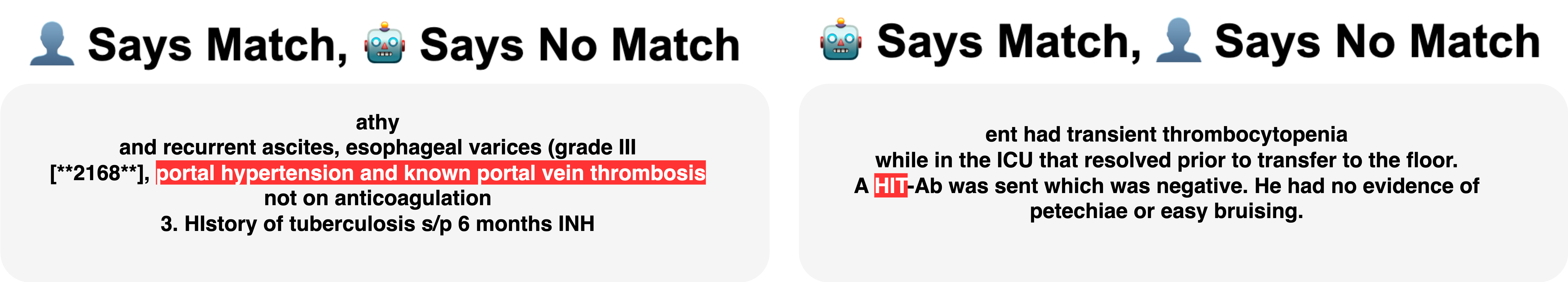}
    \caption{\textbf{Failure Modes of RDMA in Rare Disease Mining.} When reviewing existing annotations, we showcase two cases where our human annotator disagrees with RDMA's refinement step. Specifically, we see that our refinement agent fails to classify portal vein thrombosis as a rare disease, primarily because its Orphanet listing ``Non-cirrhotic and non-tumoral portal vein thrombosis" does not exactly match the entity ``portal vein thrombosis". On the other hand, while the agent is able to classify ``HIT" as ``heparin induced thrombocytopenia", it does not capture the context "which was negative" properly.}
    \label{fig:annotator_agreement_qualitative}
\end{figure}

\rev{We showcase a variety of cases from poor annotations from the original dataset in Fig. \ref{fig:ex_annotation} and our own error analysis of the 4th dataset refinement step in RDMA in Fig. \ref{fig:annotator_agreement_qualitative}.}

\begin{figure}[h!]
    \centering
    \includegraphics[width=1.0\textwidth]{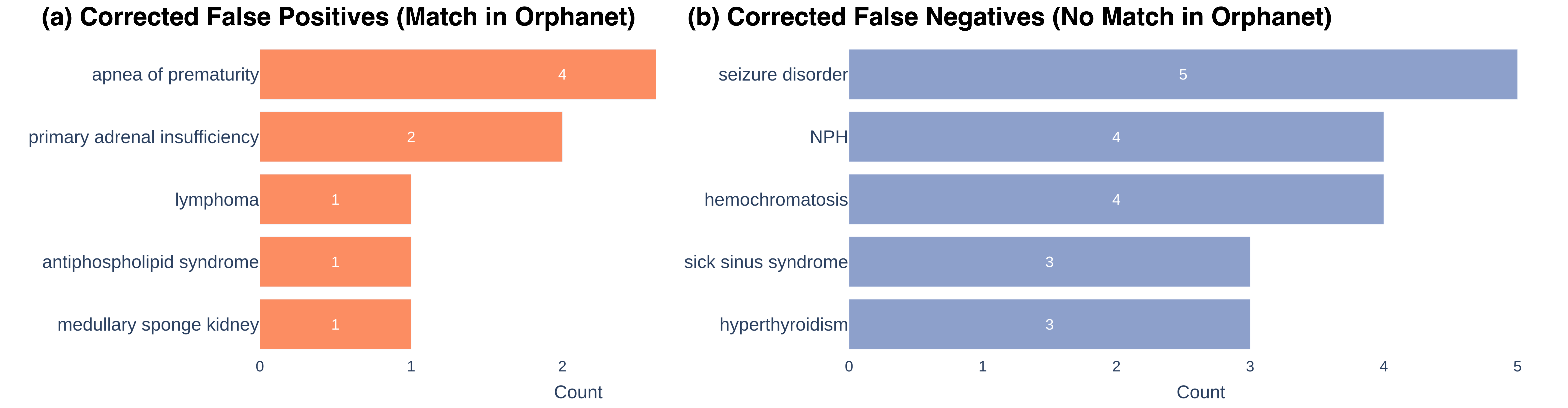}
    \caption{\textbf{Top 5 Corrected Annotations with RDMA.} RDMA effectively assists human annotation by identifying errors in the noisy historical dataset from \cite{dong2023ontology_poor_annotations}. The figure highlights the top 5 corrections, including both false negatives (a) such as seizure disorder incorrectly labeled as a rare disease in Orphanet, and false positives (b) that were missed in the initial annotations. Of the 72 annotations flagged by RDMA for human review, 55 (76.4\%) were correctly identified as erroneous, demonstrating its value as an annotation assistant for prioritizing human supervision. While hemochromatosis is typically considered a rare condition, in ICU contexts such as MIMIC notes, our physician deemed that to not necessarily be the case. Furthermore, Orphanet \cite{weinreich2008orphanet} makes a distinction between \textbf{rare} hemochromatosis and hemochromatosis. }
    \label{fig:Top_Corrections}
\end{figure}

\textbf{Examples of RDMA Flagged Entities in Dataset Refinement.} \label{appendix: dataset refinement}
We investigate which entities were primarily flagged by RDMA for human review. We visualize the top cases in Fig. \ref{fig:Top_Corrections}, where we observe that many entities mined by \cite{dong2023ontology_poor_annotations} were indeed not rare diseases, while several key rare disease entities were missed in their original annotations.

\clearpage
\section{Annotator Guidelines} \label{appendix:annotator guidelines}
\begin{figure}[h!]
    \centering
    \includegraphics[width=0.8\textwidth]{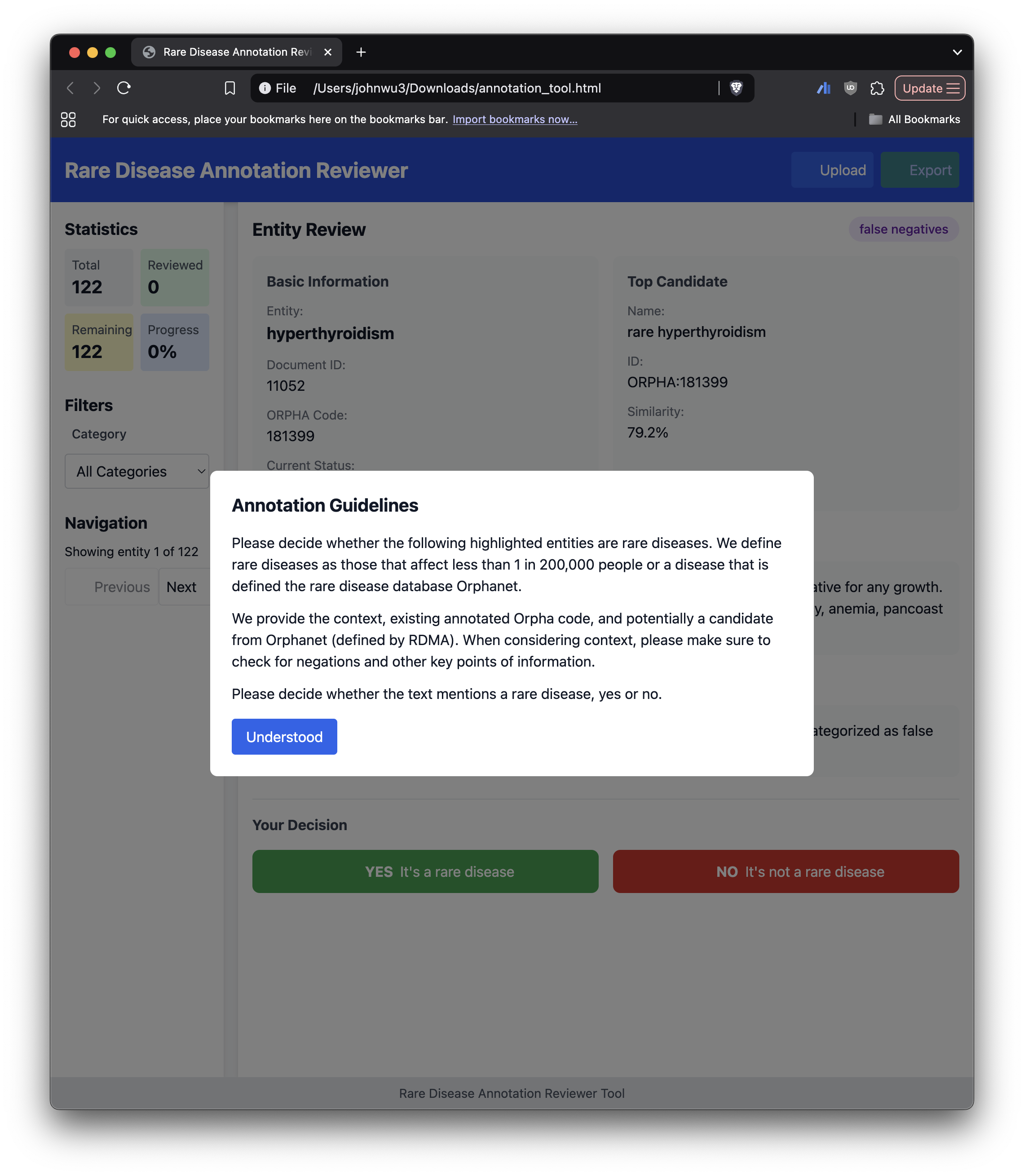}
    \caption{\textbf{Annotator Guidelines.} Annotators are asked if the mention in text directly or indirectly implies a rare disease.}
    \label{fig:annotator_guidelines}
\end{figure}

\rev{We build on the rare disease mention annotations from \cite{dong2023ontology_poor_annotations} in MIMIC-III, first filtering out clearly erroneous annotations as detailed in Table \ref{tab:Filtering_Details_ex}. Following filtering, a Ph.D. student reviewed each remaining entity from \cite{dong2023ontology_poor_annotations} under physician guidance, applying the review process outlined in Fig. \ref{fig:annotator_guidelines} to determine whether each mention directly or indirectly implies a rare disease. This produced our final curated annotation set.}

\rev{For the RDMA-assisted annotation run, the same Ph.D. student conducted a separate review pass over only the entities flagged by RDMA for human review, applying a binary correction, marking each as correct or incorrect. RDMA flags entities exhibiting disagreement across two independent mining passes and a final agent verification check, meaning the annotator is presented only with the subset of entities where automated agreement could not be established.}

\rev{\paragraph{Inter-Annotator Agreement.} We measure agreement between the human annotator and RDMA using Cohen's $\kappa$, which corrects observed agreement by chance agreement:}
\begin{equation}
    \rev{\kappa = \frac{p_o - p_e}{1 - p_e}}
\end{equation}
\rev{Given a $2 \times 2$ confusion matrix over $N = \text{TP} + \text{TN} + \text{FP} + \text{FN}$ binary judgments, observed agreement and chance agreement are computed as:}
\begin{align}
    \rev{p_o} &\rev{= \frac{\text{TP} + \text{TN}}{N}} \\
    \rev{p_e} &\rev{= p_{\text{human}} \cdot p_{\text{RDMA}} + (1 - p_{\text{human}}) \cdot (1 - p_{\text{RDMA}})}
\end{align}
\rev{where $p_{\text{human}} = (\text{TP} + \text{FN}) / N$ and $p_{\text{RDMA}} = (\text{TP} + \text{FP}) / N$ are the marginal rates of each rater labeling an entity as a rare disease mention. We report $\kappa$ under the Landis \& Koch interpretation bands, where $\kappa \in [0.41, 0.60]$ indicates moderate agreement and $\kappa \in [0.81, 1.00]$ indicates almost perfect agreement. Agreement between the human annotator and the unassisted prior annotations yielded $\kappa = 0.454$ (moderate), while agreement between the human annotator and RDMA-assisted annotations yielded $\kappa = 0.810$ (almost perfect), demonstrating a substantial improvement in annotation quality under the RDMA-assisted workflow.}

\section{Sentence Agglomeration for Faster Entity Extraction} \label{appendix: sentence agglomeration}

To optimize the performance of our retrieval-enhanced entity extraction process, we implemented a sentence agglomeration strategy that combines shorter sentences to reduce computational costs.

\begin{algorithm}
\caption{Sentence Agglomeration Algorithm}
\begin{algorithmic}[1]
\Procedure{MergeSmallSentences}{$sentences, min\_size$}
    \If{$sentences$ is empty}
        \State \Return empty list
    \EndIf
    \If{$min\_size$ is null or $min\_size \leq 0$}
        \State \Return $sentences$ unmodified
    \EndIf
    
    \State $merged\_sentences \gets$ empty list
    \State $current\_idx \gets 0$
    
    \While{$current\_idx < |sentences|$}
        \State $current\_sentence \gets sentences[current\_idx]$
        
        \If{$|current\_sentence| \geq min\_size$}
            \State Append $current\_sentence$ to $merged\_sentences$
            \State $current\_idx \gets current\_idx + 1$
            \State \textbf{continue}
        \EndIf
        
        \State $merged\_chunk \gets current\_sentence$
        \State $next\_idx \gets current\_idx + 1$
        
        \While{$next\_idx < |sentences|$ \textbf{and} $|merged\_chunk| < min\_size$}
            \If{$merged\_chunk$ is not empty \textbf{and} $sentences[next\_idx]$ is not empty}
                \State $merged\_chunk \gets merged\_chunk + $ " " $+ sentences[next\_idx]$
            \Else
                \State $merged\_chunk \gets merged\_chunk + sentences[next\_idx]$
            \EndIf
            \State $next\_idx \gets next\_idx + 1$
        \EndWhile
        
        \State Append $merged\_chunk$ to $merged\_sentences$
        \State $current\_idx \gets next\_idx$
    \EndWhile
    
    \State \Return $merged\_sentences$
\EndProcedure
\end{algorithmic}
\end{algorithm}

\begin{table}[h]
\centering

\begin{tabular}{l|c|c}
\hline
\textbf{Metric} & \textbf{No Agglomeration} & \textbf{With Agglomeration (Min Size = 500)} \\
\hline
Total documents analyzed & 117 & 117 \\
Average word count per document & 1897 & 1897 \\
Precision & 0.89 & 0.82 \\
Recall & 0.44 & 0.42 \\
F1 score & 0.59 & 0.55 \\
Extraction run time (hours) & 8:36 & 2:09 \\
\hline
\end{tabular}
\caption{\textbf{Comparative Performance Metrics for Different Sized Text Chunks in RDMA for the Entiy Extraction Step.}  We observe that performance declines slightly, but we are able to extract entities at a substantially higher rate (4x faster) from over 200,000 words.}
\end{table}

Sentence agglomeration with a minimum size of 500 characters reduced processing time by 75\% with a modest trade-off in extraction quality, making this approach particularly valuable for time-sensitive applications or large document collections.
\clearpage
\section{MIMIC3-RD Preliminary Filtering}
\textbf{Preliminary Filtering.} Before applying our RDMA refinement (Fig. \ref{fig:ex_annotation}), we filter out known incorrect keywords and add known rare disease mentions, as detailed in Table \ref{tab:Filtering_Details_ex}. These changes significantly impact the dataset statistics (Table \ref{tab:filtering_results_ex}), reducing annotations from 1,073 to 333. We use this filtered dataset for our case study in Section \ref{sec: improving existing annotations}.

\begin{table}[h]
\centering
\begin{tabular}{|p{0.3\textwidth}|p{0.7\textwidth}|}
\hline
\textbf{Kept Abbreviations} & HIT, ALS, and NPH \\
\hline
\textbf{Manually Removed Terms} & Hyperlipidemia, dyslipidemia, hypercholesterolemia \\
\hline
\textbf{Manually Added Terms} & papillary carcinoma, glioblastoma multiforme, transitional cell carcinoma, multifocal atrial tachycardia (mat), sarcoidosis, methemoglobinemia, central nervous system and systemic lymphoma, sclerosis cholangitis, mediastinitis, mesenteric vein thrombosis, multiple myeloma, hepatocellular carcinoma, primary cns lymphoma, sclerosing cholangitis, bechet's disease, neovascular glaucoma, meningocele, alopecia, neovascular glaucoma angle closure, pyoderma gangrenosum, budd-chiari, intraductal papillary mucinous tumor, complex tracheal stenosis, cervical stenosis, bronchiectasis, medullary sponge kidney, protein s, antiphospholipid antibody syndrome, protein c, hepatocellular ca, acute myelogenous leukemia, anaplastic thyroid carcinoma, thymoma, congenital bleeding disorder, tracheal stenosis \\
\hline
\end{tabular}
\caption{\textbf{Initial Keyword-based Filtering Attempts} Our clinician identified numerous incorrectly annotated terms from the original entity set. For instance, "MR" was incorrectly tagged as "Multicentric reticulohistiocytosis" in \cite{dong2023ontology_poor_annotations}'s annotations, though it commonly refers to "magnetic resonance". Our clinician flagged all abbreviations except HIT, ALS, and NPH. We also removed common disease terms like hyperlipidemia, as their rare variants are specifically defined differently in the Orphanet ontology. Finally, our clinician manually added the terms listed above. For each added term, we checked its presence in each document and, if found, added an annotation with its corresponding ORPHA code.}
\label{tab:Filtering_Details_ex}
\end{table}

\begin{table}[h]
\centering
\begin{tabular}{l c c}
\hline
\textbf{} & \textbf{Original} & \textbf{After Initial Filtering} \\
\hline
Number of Notes & 312 & 117 \\
Number of Annotations & 1,073 & 333 \\
\hline
\end{tabular}
\caption{\textbf{Rare Disease Annotation Statistics Before and After Initial Keyword Filtering.} We eliminated over 700 misannotated terms before conducting evaluations in Section \ref{sec: improving existing annotations}.}
\label{tab:filtering_results_ex}
\end{table}
\clearpage




\section{Explicit Methodology} \label{appendix: exp method}
\rev{Below, we define RDMA in substantially more detail.}

\textbf{Problem Formulation.} Let $X \coloneqq \{x_1, x_2, \ldots, x_n\}$ be a set of clinical documents, where $X$ is the corpus, $n$ is the total number of documents, and $|X| = n$. Let $P = \{p_1, p_2, \ldots, p_m\}$ where $m$ is the total number of phenotypes within the Human Phenotype Ontology (HPO) \cite{gargano2024mode_hpo} and each $p_i$ represents a specific HPO code. Let $R = \{r_1, r_2, \ldots, r_k\}$ where $k$ is the total number of rare diseases within the Orphanet ontology \cite{weinreich2008orphanet} and each $r_j$ represents a specific ORPHA code. 

Our framework, RDMA, effectively implements two key extraction functions: 
\begin{align}
\Phi_P: X \rightarrow 2^P
\end{align}
that maps each document to its set of phenotypes, where $\Phi_P(x_i) = \{p \in P \mid p \text{ is mentioned in document } x_i\}$, and 
\begin{align}
\Phi_R: X \rightarrow 2^R
\end{align}
that maps each document to its set of rare diseases, where $\Phi_R(x_i) = \{r \in R \mid r \text{ is mentioned in document } x_i\}$. The output of RDMA is an annotated dataset $D = \{(x_i, \Phi_P(x_i), \Phi_R(x_i)) \mid x_i \in X\}$ consisting of triples of documents, their mined phenotypes, and their mined rare diseases. In essence, our objective is to identify all HPO and ORPHA codes present in each clinical document.

\textbf{Tool Retrieval.} As the majority of our tools comprise databases with text content, we design them primarily as retrieval systems for LLM agent usage. 
Given a query string $q$ (e.g., a potential phenotype or disease entity) and a database of documents $\mathcal{D} = \{d_1, d_2, \ldots, d_m\}$ from an existing tool with corresponding embeddings $\mathcal{V} = \{v_1, v_2, \ldots, v_m\}$ where $v_i \in \mathbb{R}^d$, we perform similarity search to retrieve relevant documents.

The similarity between query $q$ with embedding $v_q$ and document $d_i$ with embedding $v_i$ is computed using the Euclidean distance metric:
\begin{align}
sim(q, d_i) = \frac{1}{1 + ||v_q - v_i||_2}
\end{align}

For each query $q$, we retrieve the top-$k$ documents $\mathcal{D}_q = \{d_{i_1}, d_{i_2}, \ldots, d_{i_k}\}$ such that topk$(q,\mathcal{D})$ is defined as:
\begin{align}
sim(q, d_{i_j}) \geq sim(q, d_i) \quad \forall d_i \in \mathcal{D} \setminus \mathcal{D}_q, \forall d_{i_j} \in \mathcal{D}_q
\end{align}

These retrieved documents provide contextual information necessary for entity verification and implication in subsequent stages of the RDMA framework. In our implementation, we utilize FAISS \cite{douze2025faisslibrary} for efficient indexing and searching of our document database. For embeddings, we employ MedEmbed small \cite{balachandran2024medembed} due to its optimized performance on medical tasks.

\textbf{RDMA Overview.} As illustrated in Fig. \ref{fig:RDMAvRAG}, RDMA consists of four primary steps: (1) entity extraction, (2) entity verification and implication, (3) verified entity matching, and (4) dataset refinement. We detail the prompts and setup for each step below, noting that not all steps are required for every entity, as this often depends on the specific context.

\subsection{Entity Extraction}
Each document $x_i$ can be decomposed into meaningful chunks of text such as sentences or clinical notes with standard separators like commas, periods, and other lexical symbols. Formally, we represent each document as:
\begin{align}
x_i = (s_1^i, s_2^i, \ldots, s_{l_i}^i)
\end{align}
where $s_j^i$ denotes a chunk of text (e.g., a word, sentence, multiple sentences, or pre-defined number of words) and $j \in \{1, 2, \ldots, l_i\}$ indicates the position within the document.

A critical consideration in our extraction design is the trade-off associated with chunk length selection. Excessively large chunks may dilute specific meanings, resulting in noisier retrieval of phenotype or disease candidates. Conversely, overly granular chunks significantly increase computational time for entity extraction. For our implementation, we opted for sentence-level chunking, though we explore larger sizes in Appendix \ref{appendix: sentence agglomeration} to demonstrate this trade-off.

For each sentence $s_j^i$, we retrieve the top $k=5$ rare disease or HPO candidates $C_j^i = \{c_{1j}^i, c_{2j}^i, \ldots, c_{kj}^i\}$ from the corresponding ontology using the similarity function:
\begin{align}
C_j^i = \text{topk}(s_j,\mathcal{D})
\end{align}

Both the sentence and these candidates are incorporated into our prompting strategy as illustrated in Appendix \ref{appendix:prompts}. Our objective is to extract from each document $x_i$ sets of potentially relevant entities related to phenotypes or rare diseases, which we denote as:
\begin{align}
E_{p,uv}^i &= \{e_{p,uv,1}^i, e_{p,uv,2}^i, \ldots, e_{p,uv,n_p}^i\} \\
E_{r,uv}^i &= \{e_{r,uv,1}^i, e_{r,uv,2}^i, \ldots, e_{r,uv,n_r}^i\}
\end{align}
where the subscript $uv$ indicates their "unverified" status, and the subscripts $p$ and $r$ denote phenotype and rare disease entities, respectively.

\subsection{Entity Verification and Implication} \label{sec:verify-and-implication-reasoning_ex}
We implement multiple reasoning steps for entity verification, recognizing that verification is a non-trivial process when working with clinical documents. Different verification steps are applied based on whether we are extracting phenotypes ($\Phi_P$) or rare disease mentions ($\Phi_R$). Our verification process follows these steps:

\textbf{Abbreviation detection and expansion.} First, we determine whether an extracted entity is a valid clinical abbreviation. Given an unverified entity $e_{uv}^i \in E_{p,uv}^i \cup E_{r,uv}^i$, we retrieve a set of $k=5$ abbreviation candidates $A(e_{uv}^i) = \{a_1, a_2, \ldots, a_k\}$. If $e_{uv}^i \in A(e_{uv}^i)$, we expand the term to its full form $e_{exp}^i$ and forward it to the next verification stage. Otherwise, we proceed with the original term. This process can be formalized as:
\begin{align}
e_{next}^i = 
\begin{cases}
e_{exp}^i & \text{if } e_{uv}^i \in A(e_{uv}^i) \\
e_{uv}^i & \text{otherwise}
\end{cases}
\end{align}

\textbf{Direct entity verification.} Next, we directly verify if an entity matches any entry in the HPO or Orphanet ontologies. We retrieve $k=5$ candidates, and determine whether the entity $e_{next}^i$ and its context sentence $s_j^i$ match any ontology entry. If a match exists, we mark the entity as verified ($e_{p,v}^i$ or $e_{r,v}^i$) and proceed to matching. Otherwise, for phenotype entities, we continue to lab event detection; for rare disease entities, we conclude the verification process. The verification function can be expressed as:
\begin{align}
\text{isVerified}(e_{next}^i, s_j^i) = 
\begin{cases}
\text{True} & \text{if } \exists c \in C(e_{next}^i) : \text{matches}(e_{next}^i, c, s_j^i) \\
\text{False} & \text{otherwise}
\end{cases}
\end{align}
where $C(e_{next}^i)$ represents the top-$k$ candidates from the ontology for entity $e_{next}^i$. For rare diseases, we note we also prompt the LLM if an entity is a disease, because the Orphanet ontology can contain related treatments like lab events or biological entities \cite{weinreich2008orphanet}.

\textbf{Lab event detection and implication.} For unverified implied phenotype entities, we check if they represent lab events, which often imply relevant phenotypes. We first determine if the entity contains numerical values:
\begin{align}
\text{containsNumbers}(e_{next}^i) = 
\begin{cases}
\text{True} & \text{if entity contains numerical values} \\
\text{False} & \text{otherwise}
\end{cases}
\end{align}

If numbers exist, we further validate whether the entity is indeed a lab event using an LLM-based classifier:
\begin{align}
\text{isLabEvent}(e_{next}^i) = \text{LLM\_classify}(e_{next}^i, \text{"lab event"})
\end{align}

For confirmed lab events, we retrieve the top $k=5$ lab reference ranges $L(e_{next}^i) = \{l_1, l_2, \ldots, l_k\}$ that most closely match the measured value. We then determine whether the value falls outside normal ranges:
\begin{align}
\text{isAbnormal}(e_{next}^i, L(e_{next}^i)) = \text{LLM\_reason}(e_{next}^i, L(e_{next}^i))
\end{align}

If abnormal, we generate the corresponding phenotype direction (elevated or lowered) to generate an implied phenotype entity.

\textbf{Implied phenotype generation.} If an entity is neither a direct phenotype nor a lab event, we assess whether it directly implies a phenotype:
\begin{align}
\text{impliesPhenotype}(e_{next}^i) = \text{LLM\_classify}(e_{next}^i, \text{"implies phenotype"})
\end{align}

If it does, we generate the implied phenotype:
\begin{align}
e_{p,implied}^i = \text{LLM\_generate}(e_{next}^i, \text{"implied phenotype"})
\end{align}

\textbf{Implied phenotype verification.} Finally, we verify whether the generated phenotype exists within the HPO ontology:
\begin{align}
\text{existsInOntology}(e_{p,implied}^i) = 
\begin{cases}
\text{True} & \text{if } e_{p,implied}^i \in P \\
\text{False} & \text{otherwise}
\end{cases}
\end{align}

Only verified phenotypes proceed to the matching stage.

\subsection{Verified Entity Matching}
Given the original sentence context $s_j^i$ and a verified entity $e_{p,v}^i \in E_{p,v}^i$ (for phenotypes) or $e_{r,v}^i \in E_{r,v}^i$ (for rare diseases), we match each entity to its corresponding ontological code from the top $k$ candidates. For phenotype entities, this matching process assigns HPO codes:
\begin{align}
\hat{p}_{j}^i = \text{LLM\_match}(e_{p,v}^i, s_j^i, \text{topk}(e_{p,v}^i, \mathcal{D}_{HPO}))
\end{align}
where $\hat{p}_{j}^i \in P$ represents the final predicted phenotype code and $\mathcal{D}_{HPO}$ denotes the HPO database. Similarly, for rare disease entities, we assign ORPHA codes:
\begin{align}
\hat{r}_{j}^i = \text{LLM\_match}(e_{r,v}^i, s_j^i, \text{topk}(e_{r,v}^i, \mathcal{D}_{Orphanet}))
\end{align}
where $\hat{r}_{j}^i \in R$ represents the final predicted rare disease code and $\mathcal{D}_{Orphanet}$ denotes the Orphanet database.
The complete set of predicted phenotypes and rare diseases for document $x_i$ is then:
\begin{align}
\hat{\Phi}_P(x_i) &= \{\hat{p}_{j}^i \mid \hat{p}_{j}^i \text{ extracted from sentence } s_j^i \text{ in document } x_i\} \\
\hat{\Phi}_R(x_i) &= \{\hat{r}_{j}^i \mid \hat{r}_{j}^i \text{ extracted from sentence } s_j^i \text{ in document } x_i\}
\end{align}
This matching process follows similar workflows to those in RAG-HPO frameworks \cite{garcia2024_HPORAG}, but extends the approach to rare disease identification while maintaining consistency with our established notation.

\subsection{Dataset Refinement} \label{sec: methodology - dataset refinement_ex}
Once we have assembled a set of verified and matched entities, we compare them against historically mined data, which may include previous agent-based mining attempts or human-annotated data \cite{dong2023ontology_poor_annotations} that contained flaws.

\textbf{Agentic Dataset Refinement.} Given a human or model predicted set of annotations, we re-verify existing mined entities by comparing them against newly mined annotations from RDMA. In comparing two sets of mined entities, we can compute pseudo-true positives, false positives, and false negatives.  Similar to human verification of true positives, false negatives, and false positives, our agent analyzes these categories using the verification reasoning steps defined in Section \ref{sec:verify-and-implication-reasoning}. Let $TP$, $FN$, and $FP$ represent the sets of true positives, false negatives, and false positives respectively. For each entity $e \in TP \cup FN \cup FP$, we apply the following rules:

\begin{align}
\text{flag}(e) = 
\begin{cases}
\text{no flag} & \text{if } e \in TP \text{ and isVerified}(e) \\
\text{flag} & \text{if } e \in TP \text{ and } \neg\text{isVerified}(e) \\
\text{no flag} & \text{if } e \in FN \text{ and } \neg\text{isVerified}(e) \\
\text{flag} & \text{if } e \in FN \text{ and isVerified}(e) \\
\text{flag} & \text{if } e \in FP \text{ and isVerified}(e) \\
\text{no flag} & \text{if } e \in FP \text{ and } \neg\text{isVerified}(e)
\end{cases}
\end{align}

Here, flag(e) indicates whether human expert judgment is required for a mined rare disease entity. The key idea here is that if we assume there exists noise in both mining attempts, a tertiary check is required to gauge whether or not an entity is correct. If the agent decides that a true positive or false negative, and a false positive a rare disease, then such disagreeable cases are flagged for human supervision.

\subsection{Differences in Phenotype and Disease Mining Implementation} \label{sec: Methodology: diff_pheno_dis_ex}
The document-phenotype extraction $\Phi_P$ and the document-rare disease mapping function $\Phi_R$ employ different verification steps tailored to their specific requirements, as summarized in Table \ref{tab:extraction_steps_comparison_ex}.

\begin{table}[h]
\centering
\begin{tabular}{|p{0.20\textwidth}|p{0.10\textwidth}|p{0.12\textwidth}|p{0.40\textwidth}|}
\hline
\textbf{Agentic Step} & \textbf{$\Phi_P$} & \textbf{$\Phi_R$} & \textbf{Reason} \\
\hline
Abbreviation detection & No & Yes & This is unneeded for phenotype benchmark. \\
\hline
Lab events database lookup & Yes & No & Many phenotypes are lab events. \\
\hline
Implied reasoning from context & Yes & No & Implying rare diseases is diagnosis not mining. \\
\hline
Entity Verification & HPO Check & Disease and Orphanet Check & Orphanet can contain non-disease entities. \\
\hline
\end{tabular}
\caption{Comparison of agentic steps in phenotype versus rare disease extraction. RDMA employs different agentic steps for phenotype versus rare disease extraction based on task-specific requirements.}
\label{tab:extraction_steps_comparison_ex}
\end{table}

These contrasting approaches reflect the distinct challenges inherent in each extraction task. Phenotype extraction requires inference from laboratory values and clinical observations, while rare disease extraction must carefully distinguish between common conditions and truly rare diseases while avoiding confusion with related entities such as genes, proteins, and enzymes that are also represented within the Orphanet ontology \cite{weinreich2008orphanet}. By tailoring the verification pipeline to each task's specific requirements, RDMA achieves higher accuracy without incurring unnecessary computational overhead.
\clearpage
\section{Prompts}\label{appendix:prompts}
We showcase all of the LLM prompts used in RDMA below.

\begin{figure}[h!]
    \centering
    \includegraphics[width=1.0\textwidth]{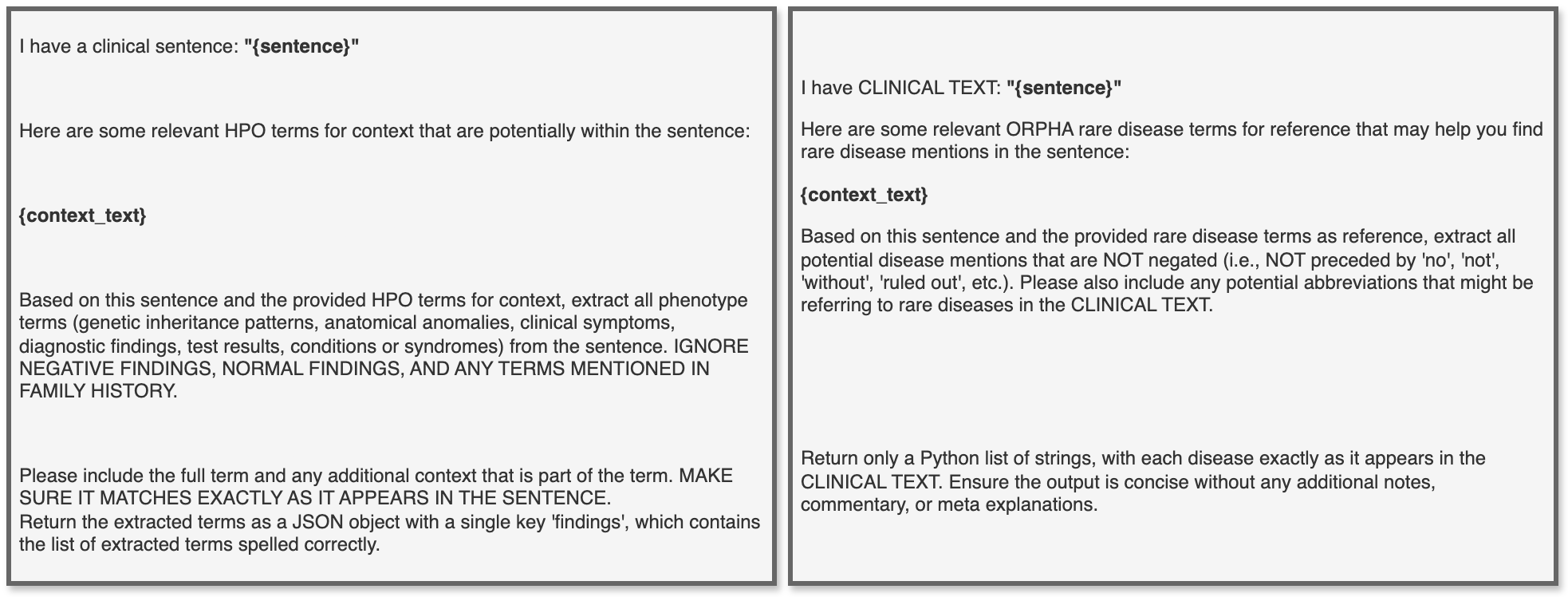}
    \caption{\textbf{Entity Extraction Prompts.} We showcase both HPO extraction (left) and Rare Disease extraction (right) prompts here.}
    \label{fig:EntityExtractionPrompt}
\end{figure}

\subsection{HPO Verification and Matching Prompts} \label{appendix: HPO prompts}
We showcase all of the prompts used for HPO extraction here. 

\begin{figure}[h!]
    \centering
    \includegraphics[width=0.8\textwidth]{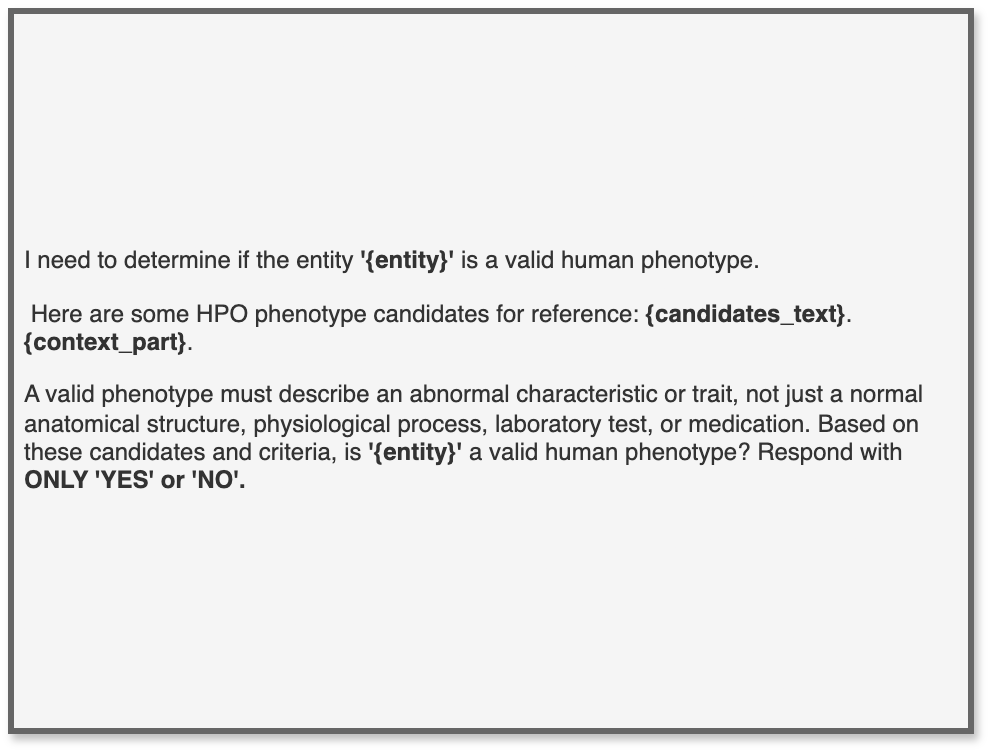}
    \caption{\textbf{Verifying an Entity is a Phenotype.} This reasoning step is used repeatedly in verifying all phenotype implications, whether done by a lab test or a generated implication. }
    \label{fig:DirectPhenotype}
\end{figure}

\begin{figure}[h!]
    \centering
    \includegraphics[width=0.8\textwidth]{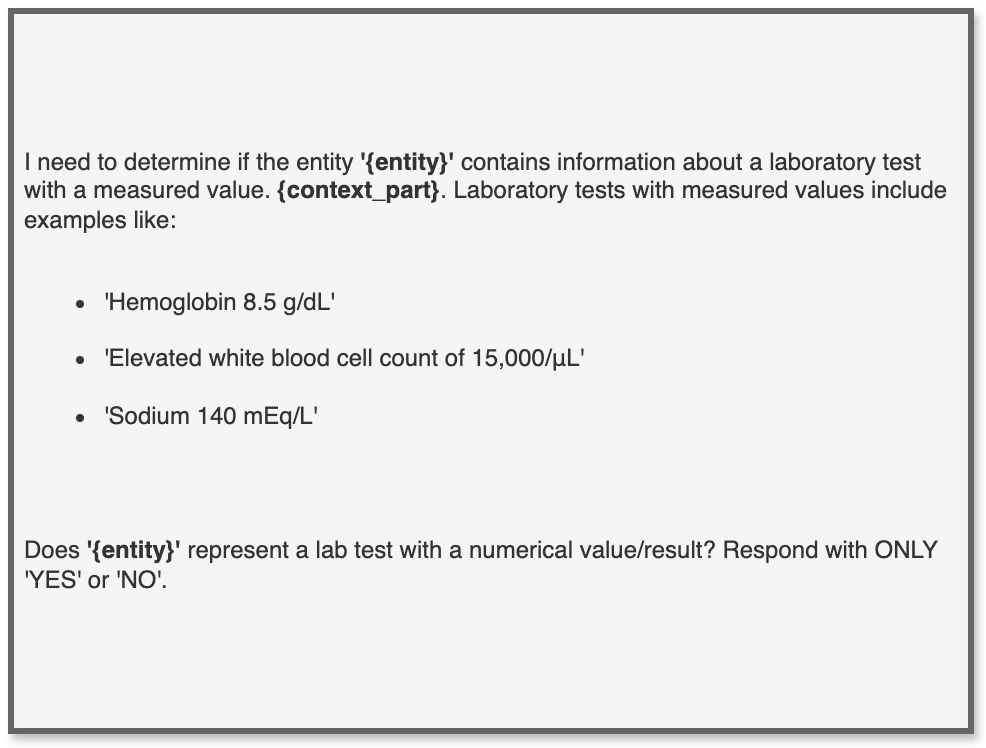}
    \caption{\textbf{Lab Test Check.} LLMs are asked to see if an entity is indicative of a lab test. }
    \label{fig:LabTestsCheck}
\end{figure}

\begin{figure}[h!]
    \centering
    \includegraphics[width=0.8\textwidth]{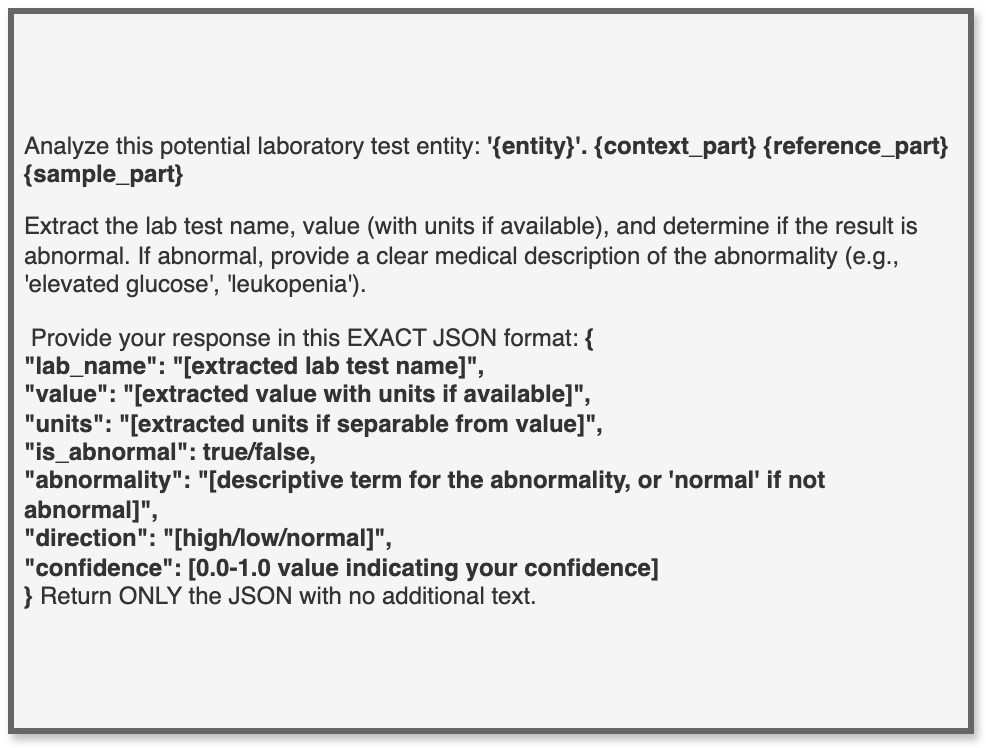}
    \caption{\textbf{Lab Test Implication.} LLMs are asked to determine if a lab event indicates an abnormality in the patient.}
    \label{fig:LabTestsAnalysis}
\end{figure}

\begin{figure}[h!]
    \centering
    \includegraphics[width=0.8\textwidth]{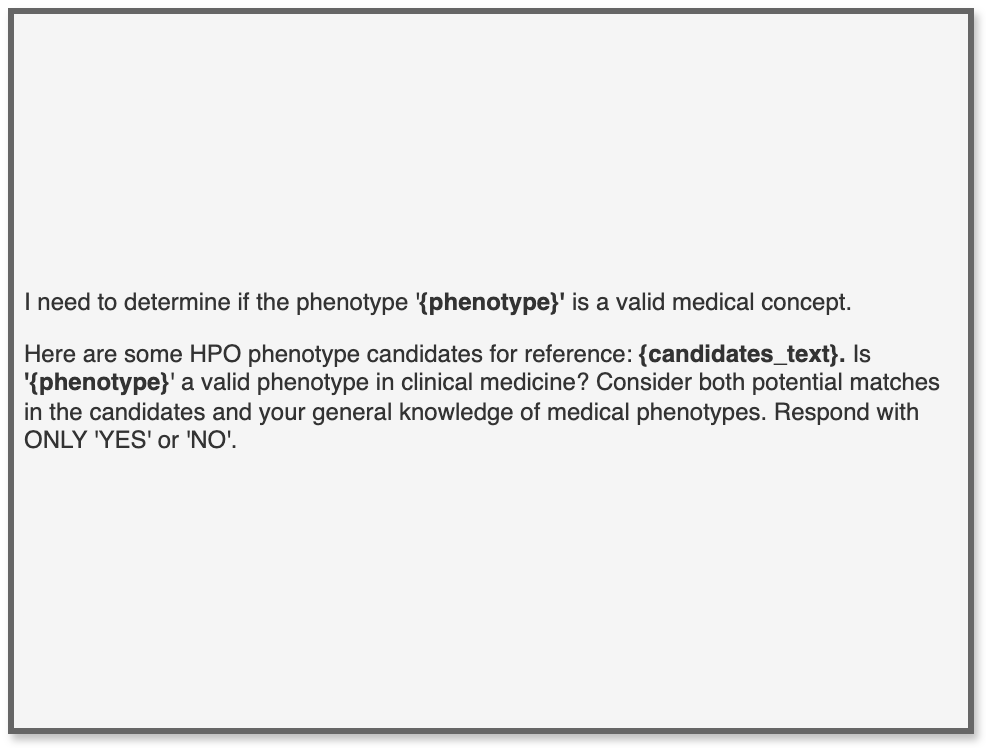}
    \caption{\textbf{Variant of Phenotype Verification.} This prompt is used to double check if an implied lab test phenotype is within the HPO ontology.}
    \label{fig:PhenotypeCorrectnessCheck}
\end{figure}

\begin{figure}[h!]
    \centering
    \includegraphics[width=0.8\textwidth]{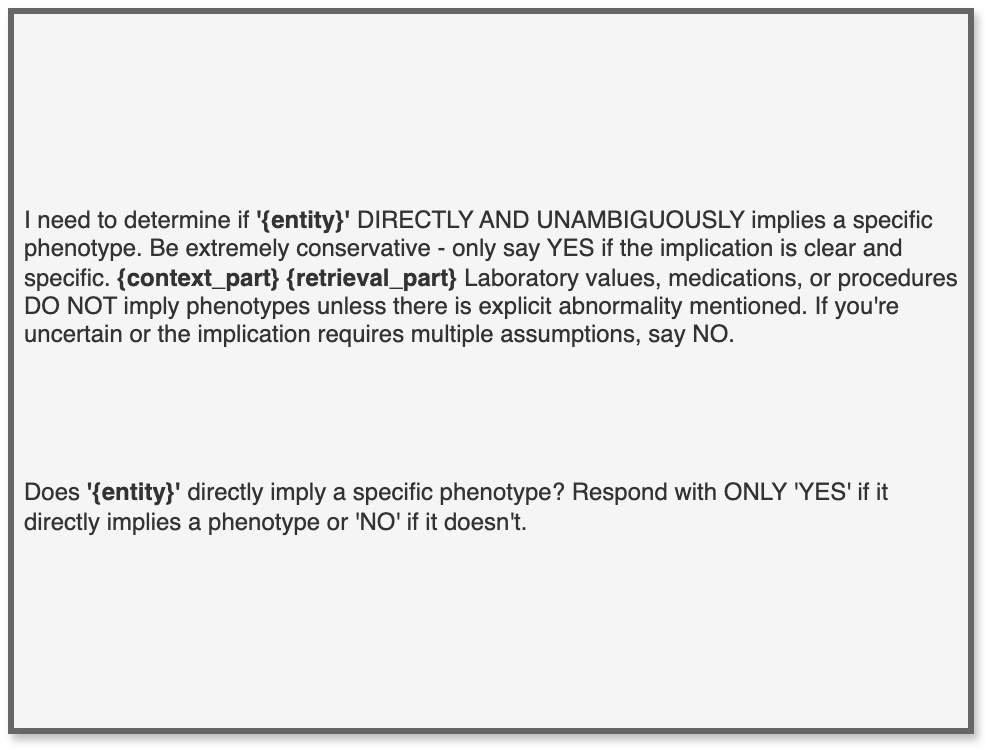}
    \caption{\textbf{Entity Implies Phenotype Check.} LLM is asked to check if an entity implies the possibility of a phenotype.}
    \label{fig:CheckIfImpliesPhenotypet}
\end{figure}

\begin{figure}[h!]
    \centering
    \includegraphics[width=0.8\textwidth]{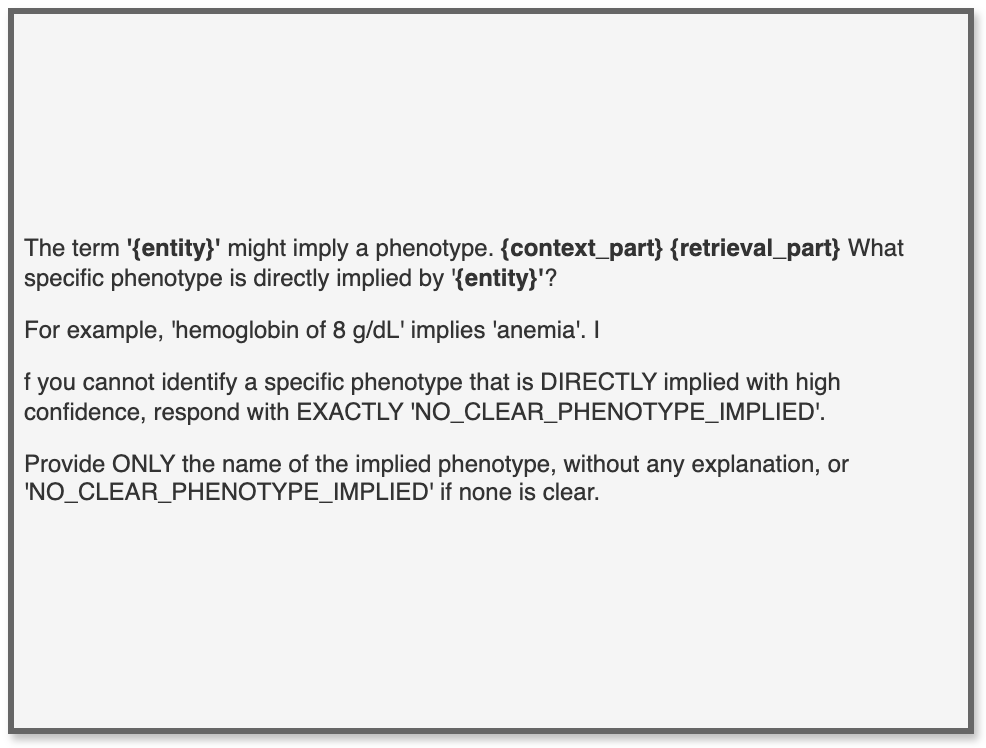}
    \caption{\textbf{Phenotype Implication Generation.} The LLM is asked to generate or predict what the implied phenotype is based on context and the entity.}
    \label{fig:PhenoImplicationGeneration}
\end{figure}

\begin{figure}[h!]
    \centering
    \includegraphics[width=0.8\textwidth]{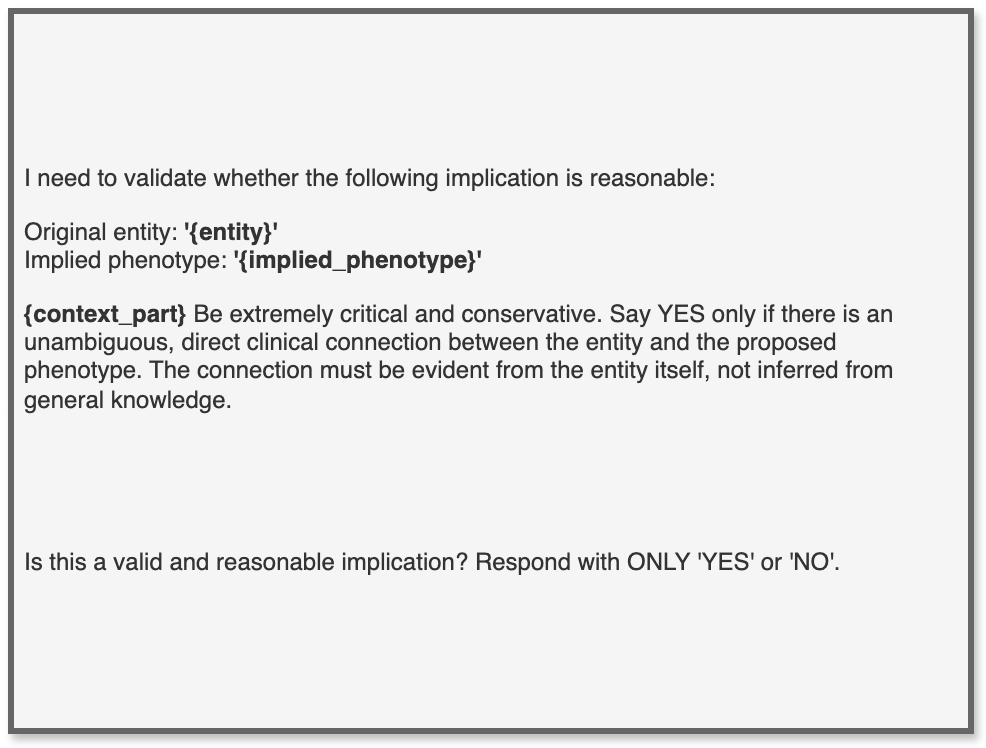}
    \caption{\textbf{Phenotype Implication Reasoning Check.} The LLM is then asked to double check its implication to make sure it makes sense.}
    \label{fig:PhenotypeImplicationReasonableCheck}
\end{figure}

\begin{figure}[h!]
    \centering
    \includegraphics[width=0.8\textwidth]{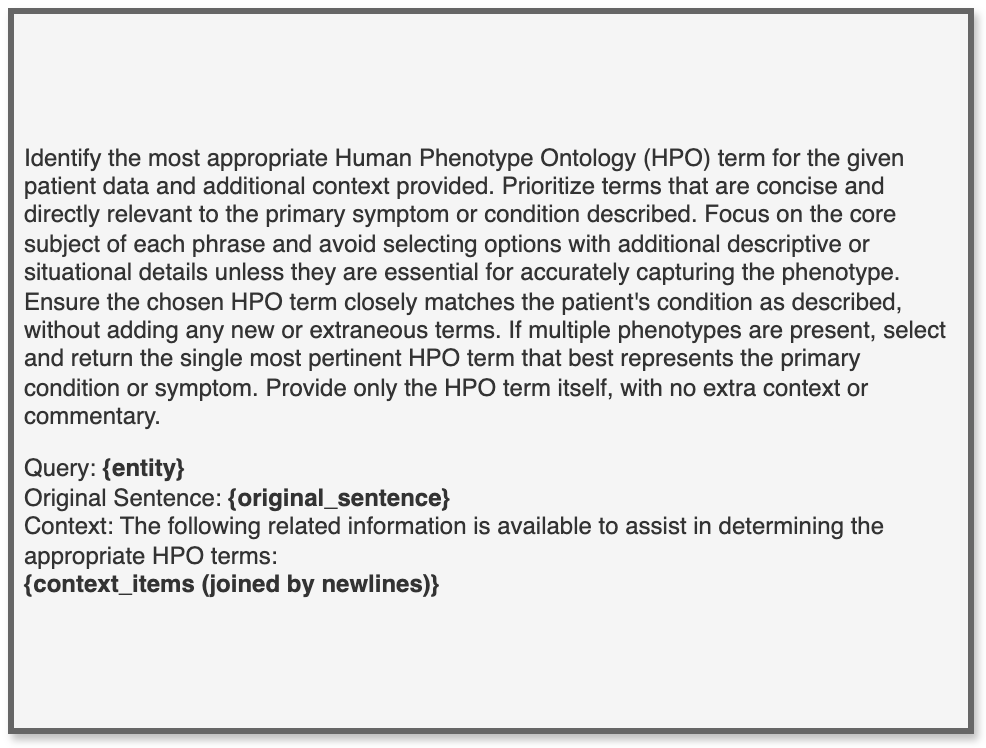}
    \caption{\textbf{Phenotype Matching.} Finally, the LLM is asked to match the an entity or implied entity to a matching phenotype.}
    \label{fig:PhenotypeMatching}
\end{figure}

\clearpage
\subsection{Rare Disease Verification and Matching Prompts} \label{appendix: rd prompts}

\begin{figure}[h!]
    \centering
    \includegraphics[width=0.8\textwidth]{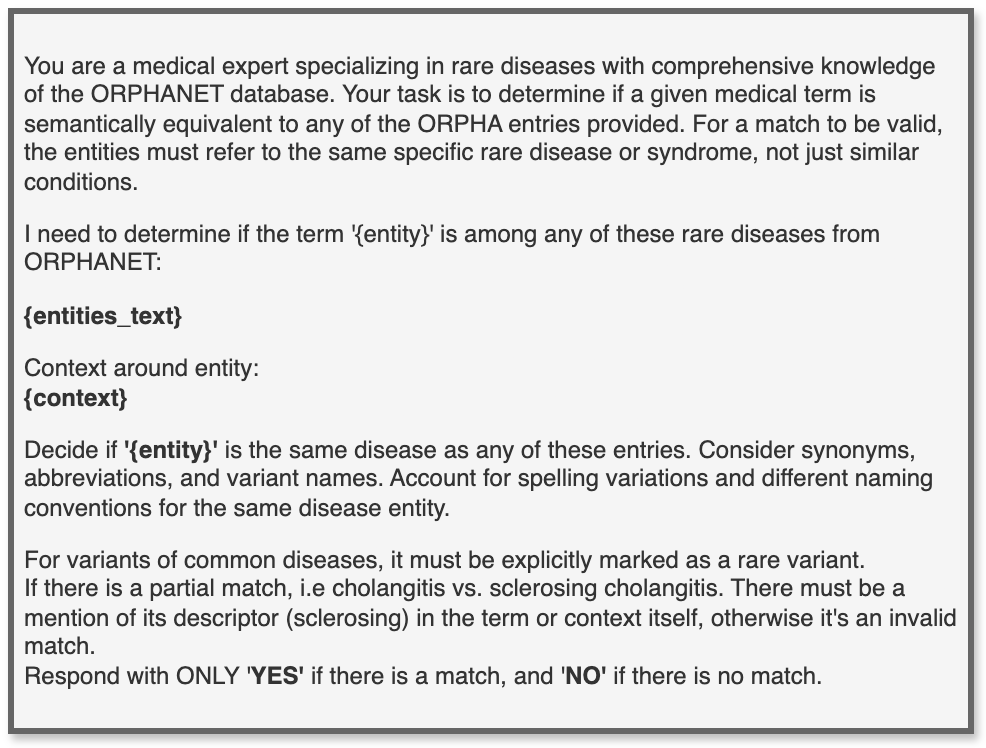}
    \caption{\textbf{Rare Disease Entity Ontology Check.} This prompt checks if the entity is within the Orphanet ontology.}
    \label{fig:RDOntologyCheck}
\end{figure}

\begin{figure}[h!]
    \centering
    \includegraphics[width=0.8\textwidth]{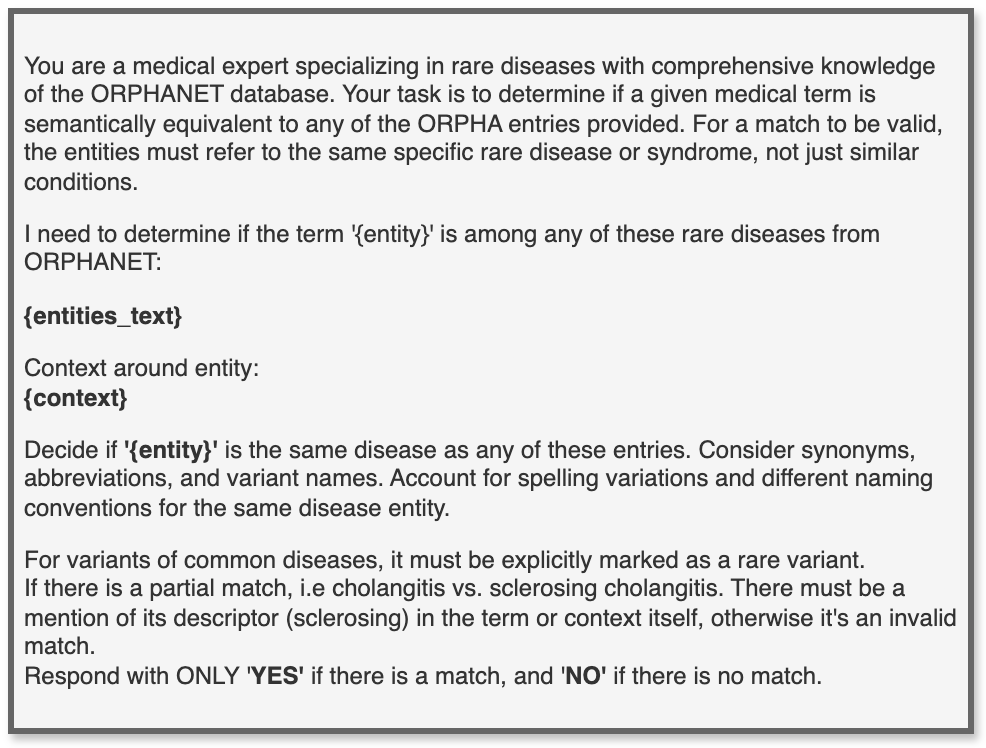}
    \caption{\textbf{Rare Disease Check.} This prompt checks if the entity is actually disease, because not all Orphanet entities are diseases. }
    \label{fig:RDDiseaseCheck}
\end{figure}

\begin{figure}[h!]
    \centering
    \includegraphics[width=0.8\textwidth]{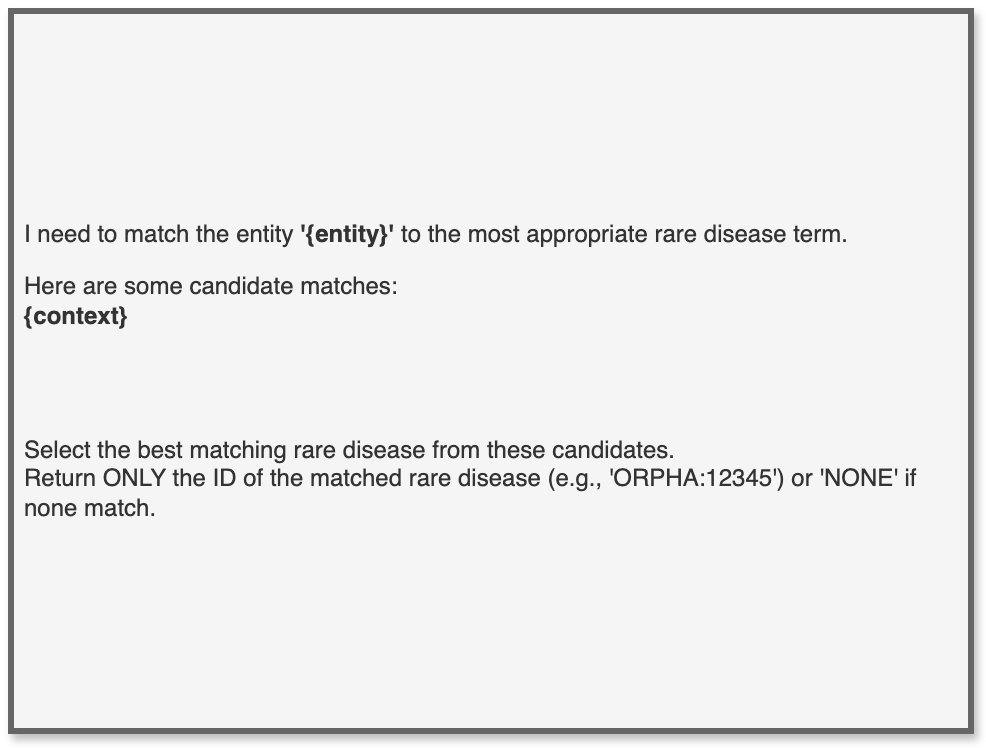}
    \caption{\textbf{Rare Disease Matching.} This prompt matches the verified entity to an Orpha code.}
    \label{fig:RDMatch}
\end{figure}

\rev{\subsection{Performance Across Annotation Sets} \label{appendix: rd annotations perf}}
\rev{We disclose what performance metrics look like across the original noisy set of annotations and our different variants of corrected annotations here.}

\begin{table}[h]
\centering
\resizebox{0.8\textwidth}{!}{%
\begin{tabular}{l c c c c}
\hline
\textbf{Baseline} & \textbf{Model} & \textbf{Precision} & \textbf{Recall} & \textbf{F1} \\
\hline
0-shot & Llama 3.3 70B$^q$ & 0.01 & 0.03 & 0.02 \\
Retrieve and String Match & MedEmbed & 0.23 & 0.36 & 0.28 \\
RAG-RD & Mistral 24B$^q$ & 0.17 & 0.45 & 0.24 \\
RDMA & Mistral 24B$^q$ & \textbf{0.67} & \textbf{0.28} & \textbf{0.39} \\
\hline
\multicolumn{5}{c}{\textbf{Human Corrected Labels}} \\
\hline
0-shot & Llama 3.3 70B$^q$ & 0.01 & 0.05 & 0.02 \\
Retrieve and String Match & MedEmbed &  0.25 & 0.43 & 0.32 \\
RAG-RD & Mistral 24B$^q$ & 0.14 & \textbf{0.54} & 0.22 \\
RDMA & Mistral 24B$^q$ & \textbf{0.66} & 0.38 & \textbf{0.48} \\
\hline
\end{tabular}%
}
\caption{\textbf{Rare Disease Extraction Performance Comparison.} We showcase performance on the other 2 sets of rare disease mention labels, the noisy original and the human only corrected labels. We observe that RDMA has significantly higher precision than its baselines, and that it overall outperforms all of its baselines in F1. \textbf{Bold} denotes best performance.}
\label{tab:label_correction}
\end{table}

\clearpage



\end{appendices}


\bibliography{sn-bibliography}

\end{document}